\documentclass[letterpaper]{article} 
\usepackage{aaai2026}
\usepackage{times}  
\usepackage{helvet}  
\usepackage{courier}  
\usepackage[hyphens]{url}  
\usepackage{graphicx} 
\usepackage{natbib}  
\usepackage{caption} 
\usepackage{algorithm}
\usepackage{algorithmic}
\usepackage{multirow}

\usepackage{newfloat}
\usepackage{listings}
\DeclareCaptionStyle{ruled}{labelfont=normalfont,labelsep=colon,strut=off} 
\floatstyle{ruled}
\newfloat{listing}{tb}{lst}{}
\floatname{listing}{Listing}
\usepackage{microtype}
\usepackage{url}
\usepackage{booktabs}
\usepackage{lineno}
\usepackage{times}
\usepackage{latexsym}
\usepackage{amsmath}
\usepackage{multirow}
\usepackage[T1]{fontenc}
\usepackage[utf8]{inputenc}
\usepackage{microtype}
\usepackage{inconsolata}
\usepackage{graphicx}
\usepackage{tikz}
\usepackage{booktabs}
\usepackage{array}
\usepackage{comment}

\definecolor{mylightblue}{RGB}{173,216,230}

\title{Confidence Estimation for Text-to-SQL in Large Language Models}
\author {
    Sepideh Entezari Maleki, 
    Mohammadreza Pourreza, 
    Davood Rafiei 
}
\affiliations {
    University of Alberta\\
    Edmonton, AB, Canada
    sentezar@ualberta.ca, pourreza@ualberta.ca, drafeiei@ualberta.ca
}

\begin{document}

\maketitle

\begin{abstract}
Confidence estimation for text-to-SQL aims to assess the reliability of model-generated SQL queries without having access to gold answers. We study this problem in the context of large language models (LLMs), where access to model weights and gradients is often constrained. We explore both black-box and white-box confidence estimation strategies, evaluating their effectiveness on cross-domain text-to-SQL benchmarks. Our evaluation highlights the superior performance of consistency-based methods among black-box models and the advantage of SQL-syntax-aware approaches for interpreting LLM logits in white-box settings. Furthermore, we show that execution-based grounding of queries provides a valuable supplementary signal, improving the effectiveness of both approaches. 
\end{abstract}


\section{Introduction}
Large Language Models (LLMs) have demonstrated remarkable proficiency in parsing natural language utterances and performing various generation tasks---such as producing text, code, and images---often at  human-like level, thereby automating a wide range of complex processes~\citep{touvron2023llama, chowdhery2022palm, brown2020language, chen2021evaluating}. One significant application of LLMs is in translating natural language queries into logic-based SQL statements, enabling non-technical users to interact seamlessly with databases \citep{rajkumar2022evaluating, gao2023text, pourreza2023din}. Despite their impressive capabilities, LLMs are not yet ready to be deployed in enterprise settings for users who lack the SQL knowledge. A major challenge is ensuring the accuracy and correctness of the generated queries. Many end-users, with limited SQL knowledge, are unable to verify if a generated query is a correct translation of their input.

A key question in this context is if the confidence level of LLMs in generated queries can be estimated, and if those estimates are accurate. Reliable confidence estimates can be quite useful. On the client side, users can evaluate generated queries alongside confidence scores, choosing to discard queries if the confidence falls below a certain threshold. On the server side, the model can abstain from generating a response when confidence is low, reducing the risk of producing incorrect or harmful outputs.
Incorporating confidence into query generation is especially important in high-stakes applications, where incorrect SQL queries can lead to significant data retrieval errors and misinformed decisions.

The issue of uncertainty in deep neural networks has been studied under various names, including out-of-distribution detection~\citep{liang2017enhancing}, uncertainty estimation\citep{lakshminarayanan2017simple}, confidence prediction~\citep{dolezal2022uncertainty}, and calibration~\citep{guo2017calibration}. Common techniques such as Monte Carlo dropout~\citep{gal2016dropout,wang2013fast} and deep ensembles~\citep{lakshminarayanan2017simple} are challenging to apply to closed-source LLMs due to limited access to model internals and training data. While open-source models offer greater transparency, challenges remain due to the computational costs and complexity of adapting these techniques at scale.
In the text-to-SQL domain, uncertainty has received limited attention, with only a few studies exploring query abstention\citep{somov2025confidence,chen2025reliable} and calibration\citep{ramachandran2024text}. More broadly, uncertainty in code generation---and text-to-SQL in particular---remains poorly understood. 
The common fallback of exhaustive testing a generated code, even with recent automation techniques~\citep{liu2024your}, is not feasible for end-users.

This paper introduces a benchmark suite for evaluating confidence estimation in text-to-SQL, along with metrics and evaluation protocols. We also propose a novel syntax-aware logit-based model that significantly improves calibration across a variety of query types and LLM families. Our work offers the first systematic comparison of black-box (e.g., output-based) and white-box (e.g., logit-based) approaches in this domain.
We evaluate both families of techniques, drawing from strategies used in LLMs more broadly~\citep{geng2024survey, zhang2023siren, wang2023survey, huang2023survey}.
Our black-box models include verbal and consistency-based approaches, incorporating various prompting strategies such as Chain-of-Thought (COT) reasoning \citep{wei2022chain, xiong2023can} and ensemble techniques that cluster multiple SQL query generations based on execution outcomes~\citep{gao2023text, sun2023sql, dong2023c3} or feature similarity~\citep{lin2023generating}.
In contrast, our white-box approaches leverage logit-based models, where token-level probabilities are aggregated using various compositional schemes to estimate the confidence of model-generated queries.

Our evaluation on Spider and BIRD, two extensive cross-domain datasets, demonstrates that white-box strategies consistently outperform black-box methods. Among the white-box methods, our SQL-syntax-aware estimation emerges as the most effective, excelling across both long and complex queries in the Bird benchmark and shorter, simpler queries in the Spider benchmark. The choice of aggregation function impacts performance, with average token probability proving to be the most stable, mitigating the impact of low-probability tokens, particularly for our SQL-syntax-aware estimation. Conversely, product aggregation, which treats token probabilities as independent and amplifies the impact of rare tokens, performs better for our schema-aware strategies in simpler query structures. Among black-box methods, the consistency-based approach, leveraging SQL query execution as a signal, is the most reliable,  though it incurs latency and cost overhead due to multiple query executions. 

We make the following key contributions:
\begin{itemize}
    \item We present the first comprehensive benchmark of confidence estimation methods for text-to-SQL, comparing black-box and white-box approaches on Spider and Bird.
    \item We introduce SAC, a syntax-aware logit aggregation method that filters out non-informative tokens and reduces expected calibration error (ECE) by up to 16\% over existing baselines
    \item We propose three white-box models tailored to different stages of SQL generation, offering interpretable and efficient confidence estimates.
    \item We adapt multiple black-box methods from other domains to text-to-SQL, enabling meaningful cross-method comparisons.
    \item Our broad evaluation across proprietary and open-source LLMs shows that white-box methods---particularly syntax-aware models using average token aggregation---consistently outperform others. Our results highlight the value of incorporating SQL structure into confidence estimation.
\end{itemize}

\section{Related Works}



Confidence prediction and uncertainty estimation have been widely explored in traditional supervised learning \citep{gawlikowski2023survey, lakshminarayanan2017simple, guo2017calibration}. The shift to LLMs introduces new challenges in confidence estimation, particularly in generative tasks. Black-box methods rely solely on model outputs, using verbalized confidence \citep{lin2022teaching}, consistency-based techniques \citep{manakul2023selfcheckgpt}, and surrogate models \citep{shrivastava2023llamas} to infer uncertainty without accessing the internal workings of the model.

White-box methods, on the other hand, utilize internal model states, such as logits or activations, to estimate confidence. Logit-based approaches measure uncertainty at the token or sentence levels \citep{huang2023look}, while semantic-based methods adjust confidence using token relevance and semantic similarity \citep{stengel2023calibrated, duan2023shifting}. Additionally, probing methods analyze model activations to classify whether an LLM ``knows'' the answer \citep{kadavath2022language, azaria2023internal}. \citet{ramachandran2024text} show that simply rescaling a model’s full‑sequence probability already calibrates text‑to‑SQL better than self‑check prompts. 

A related line of work explores query abstention, where the system opts not to produce a query when confidence is low. TriageSQL \citep{zhang2020did} and TrustSQL \citep{lee2024trustsql} introduce benchmarks for query abstention but lacked methods for direct confidence scoring. \citet{somov2025confidence} treat confidence as a  thresholding mechanism over  token entropy or  sequence likelihood. \citet{chen2025reliable} presents RTS, a method that abstains during schema linking with conformal guarantees and optional human intervention.

Our work unifies and extends these directions. We systematically compare seven black-box and white-box confidence estimation methods for text‑to‑SQL, evaluate them on two benchmarks, and introduce a syntax-aware logit aggregation method that ignores low‑signal tokens and focuses on schema‑linked elements. Unlike prior abstention-based methods, we provide calibrated scalar confidence scores rather than binary decisions.

\begin{figure*}[h]
  \centering
  \includegraphics[width=0.7\textwidth]{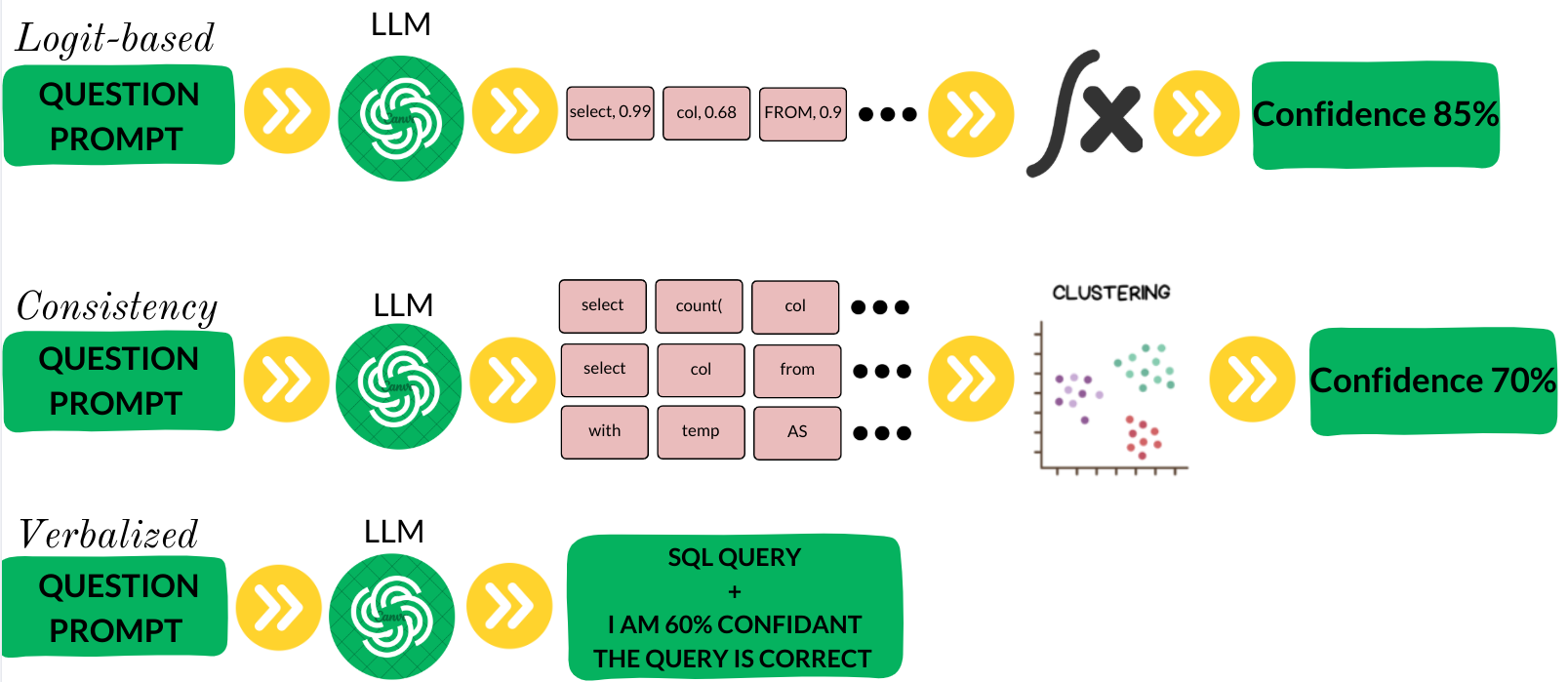}
  \caption{Overview of our methods: verbalized, consistency-based, and logit-based confidence prediction}
\label{fig:1}
\end{figure*}

\section{Methodology}
\label{sec:3}
Our approach to confidence prediction encompasses three strategies: verbalized, consistency-based, and logit-based methods, as illustrated in Figure~\ref{fig:1}.
Verbalized confidence prediction entails directly instructing the LLM to articulate its confidence in the generated text, whereas our consistency-based approach  generates multiple responses and employs the consistency among these responses as an indicator of confidence. 
Our logit-based approach  leverages the principle that an LLM's token generation is guided by a sequence of probability distributions (logits) over the vocabulary, where the most likely next token is predicted at each step based on the input context.

\subsection{Verbalized Confidence}
\label{sec:3.1}
Verbalized confidence prediction encompasses a range of approaches where the model is prompted to generate a confidence score alongside its response to a given question. 
This approach mirrors the way we seek uncertainty assessments from human experts, by asking them to express their confidence in their recommendations. 
We investigate three variations of verbalized confidence: vanilla verbalized, chain-of-thought, and augmented chain-of-thought. Each variation incrementally increases the information included in the prompts, allowing us to assess how LLMs respond to different levels of contextual detail. 

\paragraph{Vanilla verbalized}
The most direct approach involves instructing the model to generate a confidence score within a range of 0 to 100 \citep{xiong2023can, manakul2023selfcheckgpt}. This approach employs an input prompt consisting of three key elements: an instruction to guide the model in generating both a SQL query and a corresponding confidence score, the question presented to the model, and the database schema, complete with three sample rows for each table. Including the database schema and sample rows in the prompt serves the purpose of aiding the model in crafting a SQL query, a practice commonly observed in the literature \citep{rajkumar2022evaluating}. Detailed prompts are provided in Appendix.

\paragraph{Chain-Of-Thought verbalized}
The prediction of confidence levels demands a high degree of reasoning ability, involving an assessment of the model's certainty regarding its generated responses. This entails the model not only comprehending the context, represented by the database schema, and the content of the input question but also engaging in metacognitive reasoning to gauge the accuracy of its output. Building on insights from previous work \citep{xiong2023can}, to facilitate more effective reasoning, we employ a one-shot \textit{chain-of-thought} prompting technique for confidence prediction, as introduced in \citet{kojima2022large, wei2022chain}. The input prompt for this method includes instructions, the given question, and the database schema with sample rows, similar to the vanilla confidence prediction. Additionally, we incorporate a reasoning example, commencing with the phrase: ``Let's think step by step.'' Detailed prompts can be found in the appendix section.

\paragraph{Augmented chain-of-thought}
In text-to-SQL tasks, relying solely on the SQL query, database schema, and the given question may be insufficient for accurate confidence prediction. An additional source of valuable information, unique to text-to-SQL, is the execution result of the SQL query, which is unavailable in other areas like question answering. We improve the chain-of-thought confidence prediction method by incorporating the execution results of the generated SQL query. This augmentation provides the model with additional context, enabling more precise confidence scoring.
To maintain efficiency and avoid exceeding the model’s context window, we constrain the execution result to the first 1000 distinct rows. The prompt for this method builds upon the Chain-of-Thought approach, incorporating execution results directly before the reasoning process.

\paragraph{Self-Check Confidence}
Inspired by the findings of \citet{kadavath2022language} and \citet{ramachandran2024text}, which show that language models can self-assess the truthfulness of their outputs, we adopt a verbalized self-evaluation strategy with two variants: (T) indicating that the query is correct, and (F) indicating that it is incorrect.

\subsection{Consistency-based Confidence}
\label{sec:3.2}

Influenced by the self-consistency method, a common approach for improving text-to-SQL generation involves producing multiple output samples and selecting the most consistent answer as the final prediction \citep{gao2023text, sun2023sql, dong2023c3}. In our study, we leverage the consistency between generated SQL queries as a metric for confidence estimation. To produce multiple samples for a given question, we employ the same technique as proposed by the self-consistency method, increasing the temperature to introduce randomness during SQL generation. Raising the temperature redistributes some probability mass from highly likely tokens to less likely ones. However, if the initial probability is already sufficiently high, the token should still have a greater chance of being chosen as the next token. Conversely, when the model lacks confidence in its response, increasing the temperature can make the probability distribution approach a more uniform distribution. As a result, sampling multiple times from the model can yield significantly different answers. Next, we explore various methods for quantifying the consistency between generated samples.

\paragraph{Execution-based consistency}
SQL query execution on a database instance provides definitive answers to a given question, making execution results a valuable basis for consistency analysis. In our approach, we generate multiple SQL query samples from the LLM, execute them on the database, and cluster these queries based on the extent of exact matches in their retrieved rows. The confidence score for each SQL query is then computed as the proportion of model-generated queries that produce the same execution result, reflecting the model's certainty in its predictions.

\paragraph{Embedding-based consistency}
While execution-based consistency proves effective in predicting confidence scores, it may not be suitable for large-scale enterprise databases due to the computational cost of executing multiple SQL queries and comparing results. To address this limitation, our embedding-based approach generates contextual embeddings for SQL queries and clusters these embeddings to identify consistencies among the samples. Similar to the execution-based method, the confidence score for a query sample is estimated as the fraction of queries that fall in the same cluster.

\paragraph{Schema-based consistency}
Several studies have demonstrated the pivotal role of schema links---including table names, column names and mentioned constants---in ensuring the correctness of SQL queries generated by LLMs \citep{pourreza2023din, li2023can}. In many cases, the accuracy of schema link selection outweighs the specific arrangement of these links with SQL keywords for SQL query correctness.
Building on this insight~\citep{pourreza2023din, cao2021lgesql, wang2019rat, guo2019towards}, our schema-based confidence prediction identifies all schema links from generated SQL query samples and evaluates their consistency by performing an exact match of these schema links across the samples. This method is supported by the observation that multiple correct SQL queries can often be generated for a given question, with the shared schema links serving as a common feature. SQL queries with identical schema links are grouped into clusters, and the confidence score for each query is calculated as the proportion of queries within its cluster.

\subsection{Logit-based Confidence}
\label{sec:logit-based}

Many LLMs, including both open-source and proprietary models like GPT-4o, output token logits that represent their confidence in predicting the next tokens.
These logits are influenced by the LLM's inherent understanding of SQL syntax and the input prompt, 
 which includes the SQL schema, query, and relevant examples or context. Our logit-based approach utilizes these logits for confidence prediction. The input prompt for this method mirrors the structure of the vanilla verbal model, but the output includes both the generated tokens and their associated probabilities.

\paragraph{Full-token confidence}
This method treats all tokens in a generated SQL query---including SQL keywords, table and column names, operators, functions, and formatting elements---equally weighted, aggregating their generation probabilities.

\paragraph{Schema-Linked confidence}
This method builds on our schema-based approach discussed under consistency-based models, emphasizing the critical importance of accurate schema mapping for query correctness~\citep{pourreza2023din}. 
 It exclusively focuses on schema-linked tokens, such as table and column names and condition values, to estimate the model confidence.

\paragraph{SQL-Aware confidence}
Recognizing that not all components of an SQL query hold equal significance, this method refines confidence estimation by incorporating SQL-specific conventions and distinguishing between critical and non-critical tokens. The goal is to ensure that confidence scores reflect the semantic correctness of queries by prioritizing key tokens such as Schema-Linking, JOIN conditions, and WHERE clauses ~\citep{pourreza2023din} while normalizing or ignoring less important elements, such as formatting tokens and optional keywords. This refinement is achieved through a few steps, as discussed in the next section (see Appendix for more detailed samples).

\textit{Token Exclusion}
Certain tokens---including extra whitespaces, redundant parentheses, and optional keywords such as \texttt{INNER} in \texttt{INNER JOIN} or \texttt{AS} used for aliasing---are excluded from consideration as they do not impact the correctness of an SQL query. This allows the focus to remain on elements critical to query accuracy.

\textit{Case Folding and Order Folding} SQL syntax is insensitive to the casing of reserved keywords (e.g., \texttt{SELECT}, \texttt{FROM}, \texttt{WHERE}) and the references to table and column names. To account for this, probabilities for case variants (e.g., \texttt{select} and \texttt{SELECT}) are combined, treating them as equivalent. Similarly, the order of elements within clauses such as \texttt{SELECT}, \texttt{FROM}, \texttt{WHERE}, and \texttt{GROUP BY} does not affect query correctness. For instance, \texttt{SELECT a, b} and \texttt{SELECT b, a} are  equivalent. Where applicable and safe, probabilities are adjusted for this flexibility, ensuring the accuracy of confidence estimation, see Appendix.

\textit{Synonym Folding} Synonymous keywords (e.g., \texttt{not equal to}, \texttt{!=}, \texttt{<>}), symmetric conditions (e.g., \texttt{x=y} and \texttt{y=x}), and logically equivalent expressions (e.g., \texttt{a AND b} and \texttt{b AND a}) are treated as interchangeable. Probabilities are adjusted accordingly, where safe, to reflect this equivalence. For a detailed explanation of the probability adjustments for case insensitivity, synonymous keywords, and interchangeable constructs, see Appendix.

\paragraph{Aggregation of token probabilities}
Our logit-based models aggregate token probabilities into a single confidence score that reflects the model's overall certainty for the generated SQL query. Assuming each token generation is an independent event, the confidence in the query can be calculated as the product of the individual token probabilities. 
This method, referred to as the \textit{product method}, emphasizes the joint confidence across all tokens and is highly sensitive to any low-probability token, making it particularly effective for short and schema-heavy queries where every token is critical. However, for longer or more complex queries, this sensitivity can result in disproportionately low confidence scores due to the cumulative effect of even minor uncertainties.

As an alternative that closely resembles perplexity~\footnote{Perplexity is defined as $2^{-m}$ where m is the mean of log probabilities.}, 
the \textit{average method} calculates the mean probability across all tokens.
This method provides a normalized view of token probabilities, making it robust for longer or more complex queries. By smoothing the impact of low-probability tokens, the average method ensures a balanced evaluation. However, it may under-penalize critical tokens with very low probabilities, particularly in cases where schema-linked elements or logical conditions are essential to query correctness.

\label{sec:3.3}

\section{Experiments}
\subsection{Evaluation Setup}
\paragraph{Datasets} Our evaluation was conducted on the development sets of Spider~\citep{yu2018spider} and BIRD~\citep{li2024can}, two well-established cross-domain text-to-SQL benchmarks.
The Spider dataset consists of 1,034 examples across 20 databases, and the BIRD dataset includes 1,533 question-SQL pairs from 11 databases, covering diverse domains such as healthcare, finance, and education. The BIRD dataset presents more complex SQL query challenges compared to Spider.

\paragraph{LLMs}  We conduct our experiments using both closed-source and open-source large language models (LLMs), including GPT-3.5-turbo and GPT-4o (closed-source), as well as DeepSeek 6.7B and Qwen2.5 (open-source). 

\paragraph{Metrics}
Execution accuracy is the gold standard for text-to-SQL, but it requires ground-truth queries.  We therefore evaluate each confidence score with two standard proxies \citep{becker2024cycles,xiong2023can}: \textbf{AUC-ROC}, the area under the ROC curve, quantifies how well a score separates correct from incorrect queries (1.0 = perfect, 0.5 = random).   \textbf{Expected Calibration Error (ECE)} bins predicted probabilities and reports the mean gap between confidence and empirical accuracy (lower is better; 0 means perfect calibration) (see more details in Appendix).


\paragraph{Baselines.}
Our black-box baselines are inspired by established hallucination detection techniques for closed-source language models~\citep{manakul2023selfcheckgpt}, adapted to the text-to-SQL setting. We adopt the full-token confidence (FTC) framework from~\citet{ramachandran2024text}; Among the variants evaluated, the FTC-Product has demonstrated the best performance on both the Spider and BIRD benchmarks, making it a strong baseline for our comparisons.

\subsection{Models Compared}
\begin{table*}[!tb]
\caption{\small Performance on Spider and Bird dev sets.  
FTC = Full Token Confidence, SLC = Schema-Linked Confidence, SAC = SQL-Aware Confidence.  
\textbf{*} indicates statistically significant improvement over the best baseline (p < 0.05).}
\label{tab:main_results}
\centering
\small
\setlength{\tabcolsep}{2pt}     
\renewcommand{\arraystretch}{1}  
\begin{tabular}{l l l
                c c  
                c c  
                c c  
                c c} 
\hline
\textbf{Dataset} & \textbf{Approach} & \textbf{Method} &
\multicolumn{2}{c}{\textbf{GPT-3.5}} &
\multicolumn{2}{c}{\textbf{DeepSeek}} &
\multicolumn{2}{c}{\textbf{GPT-4o-mini}} &
\multicolumn{2}{c}{\textbf{Qwen2.5}}\\
 & & & AUC$\uparrow$ & ECE$\downarrow$ & AUC & ECE & AUC & ECE & AUC & ECE \\
\hline
\multirow{12}{*}{Spider} 
 & \multirow{6}{*}{Logit-based} 
 & FTC – Average  & 63.07 & 22.10 & 58.44 & 24.41 & 68.35 & 20.23 & 61.81 & 23.36 \\
 & & FTC – Product  & 68.10 & 19.38 & 63.04 & 20.00 & 77.23 & 17.15 & 70.54 & 19.56 \\
 & & SLC – Average  & 62.98 & 23.70 & 53.08 & 29.07 & 65.70 & 23.19 & 63.77 & 25.82 \\
 & & SLC – Product  & 65.38 & 22.13 & 61.23 & 22.67 & 69.93 & 20.94 & 67.04 & 20.70 \\
 & & SAC – Average  & 65.01 & 23.08 & 58.12 & 27.90 & 68.89 & 21.75 & 65.11 & 21.91 \\
 & & SAC – Product  & \textbf{71.66}$^{*}$ & \textbf{16.98} & \textbf{65.26}$^{*}$ & \textbf{19.80} & \textbf{79.87}$^{*}$ & \textbf{14.11} & \textbf{74.89}$^{*}$ & \textbf{15.08} \\

 \cline{2-11}
 & \multirow{6}{*}{Black-box} 
 & Consistency(Exec) & 65.91 & 19.13 & 63.25 & 19.98 & 71.82 & 17.21 & 68.32 & 19.04 \\
 & & Consistency (Embed) & 58.42 & 24.34 & 60.62 & 23.07 & 65.31 & 21.52 & 63.74 & 22.30 \\
 & & Consistency (Schema) & 59.97 & 23.04 & 61.65 & 22.34 & 67.32 & 20.11 & 65.76 & 20.78 \\
 & & Verbalized (Vanilla) & 55.45 & 25.98 & 56.04 & 25.29 & 56.14 & 25.42 & 54.91 & 26.01 \\
 & & Verbalized (COT) & 58.14 & 24.93 & 60.57 & 23.11 & 57.20 & 24.47 & 55.39 & 25.11 \\
 & & Verbalized (Aug COT) & 58.70 & 23.54 & 60.94 & 22.97 & 61.05 & 22.78 & 59.32 & 23.29 \\
  & & Self-Check Bool & 55.81 & 25.16 & 58.24 & 25.61 & 57.25 & 25.11 & 56.81 & 25.01 \\
\hline
\multirow{12}{*}{Bird} 
 & \multirow{6}{*}{Logit-based} 
 & FTC – Average  & 74.88 & 19.37 & 74.02 & 20.65 & 75.18 & 19.04 & 76.58 & 19.01 \\
 & & FTC – Product  & 72.36 & 28.21 & 67.77 & 27.34 & 72.43 & 24.04 & 72.31 & 25.11 \\
 & & SLC – Average  & 77.07 & 19.18 & 76.66 & 20.23 & 78.54 & 18.20 & 78.08 & 20.21 \\
 & & SLC – Product  & 73.71 & 23.90 & 69.88 & 23.53 & 77.73 & 20.25 & 74.03 & 22.96 \\
 & & SAC – Average  & \textbf{79.06}$^{*}$ & \textbf{12.15} & \textbf{77.94}$^{*}$ & \textbf{11.06} & \textbf{83.06}$^{*}$ & \textbf{10.03} & \textbf{79.94}$^{*}$ & \textbf{11.03} \\
 & & SAC – Product  & 73.00 & 22.88 & 71.19 & 21.11 & 81.36 & 18.23 & 75.27 & 19.19 \\
 \cline{2-11}
 & \multirow{6}{*}{Black-box} 
 & Consistency (Exec) & 67.36 & 27.14 & 66.48 & 28.34 & 76.34 & 20.74 & 71.18 & 26.11 \\
 & & Consistency (Embed) & 61.21 & 31.12 & 60.96 & 34.31 & 73.82 & 24.75 & 67.23 & 28.89 \\
 & & Consistency (Schema) & 64.47 & 29.98 & 63.16 & 30.51 & 74.14 & 23.32 & 70.08 & 27.12 \\
 & & Verbalized (Vanilla) & 55.67 & 58.61 & 56.23 & 54.22 & 58.43 & 52.02 & 56.14 & 53.22 \\
 & & Verbalized (COT) & 57.12 & 53.11 & 58.84 & 50.49 & 59.48 & 48.77 & 57.03 & 50.26 \\
 & & Verbalized (Aug COT) & 57.73 & 52.29 & 59.21 & 46.31 & 63.16 & 29.99 & 60.78 & 33.23 \\
   & & Self-Check Bool & 55.81 & 48.11 & 56.69 & 44.42 & 59.51 & 42.52 & 56.46 & 43.16 \\
\hline
\end{tabular}
\end{table*}

As summarized in Table~\ref{tab:main_results}, our logit-based methods, especially SQL-Aware Confidence (SAC) consistently outperform black-box approaches, including consistency-based, verbalized variants, and Self-check bool, \textit{across both benchmarks and all four LLMs we evaluate} (GPT-3.5, DeepSeek-6.7B, GPT-4o-mini, and Qwen2.5-Coder).
SAC delivers up to a 12\% improvement in AUC-ROC and reduces ECE by as much as 16\% on BIRD relative to the best black-box competitor (p < 0.05). BIRD’s join-heavy queries expose many schema-linked tokens that SAC leverages, whereas Spider’s shorter pattern-style queries provide fewer such cues and are therefore harder to calibrate.  By focusing on schema-linked tokens and critical SQL operators, SAC mitigates both overconfidence and underconfidence.  The two newer open-source models (GPT-4o-mini and Qwen2.5-Coder) follow the same trend: SAC-Avg remains best on BIRD, while SAC-Prod leads on Spider (p < 0.05 in both cases).

The aggregation methods handles uncertainty differently. Product aggregation is highly sensitive to low-confidence tokens, which are more common in longer or more complex queries. This sensitivity can lead to overly pessimistic scores. In contrast, average aggregation dilutes the influence of any single low-confidence token, making it more robust in challenging settings. The SAC method consistently outperforms other models on both datasets because it is robust and focuses only on the most important tokens in the SQL query—namely, schema-related tokens like table names, column names, and values. This syntax-aware focus makes it less sensitive to irrelevant variations and enhances its reliability across different scenarios.
  
Execution-based consistency stands out among black-box models, achieving the highest AUC-ROC and lowest ECE scores, highlighting its effectiveness in leveraging external feedback through query execution when model logits are unavailable. On the Spider dataset, it occasionally matches or even outperforms certain logit-based methods, demonstrating its ability to capture correctness in simpler query scenarios reliant on execution feedback. In contrast, verbalized approaches often exhibit poor calibration, with ECE exceeding 50\% in some cases, indicating significant overconfidence. The narrower performance gap between verbalized methods and logit-based methods on Spider can be attributed to Spider's simpler query structures, which improves the effectiveness of verbalized confidence estimation.





\subsection{Performance Varying Query Difficulty}
\begin{figure}[t]
\centering
\includegraphics[width=\columnwidth,trim=10 5 10 10,clip]{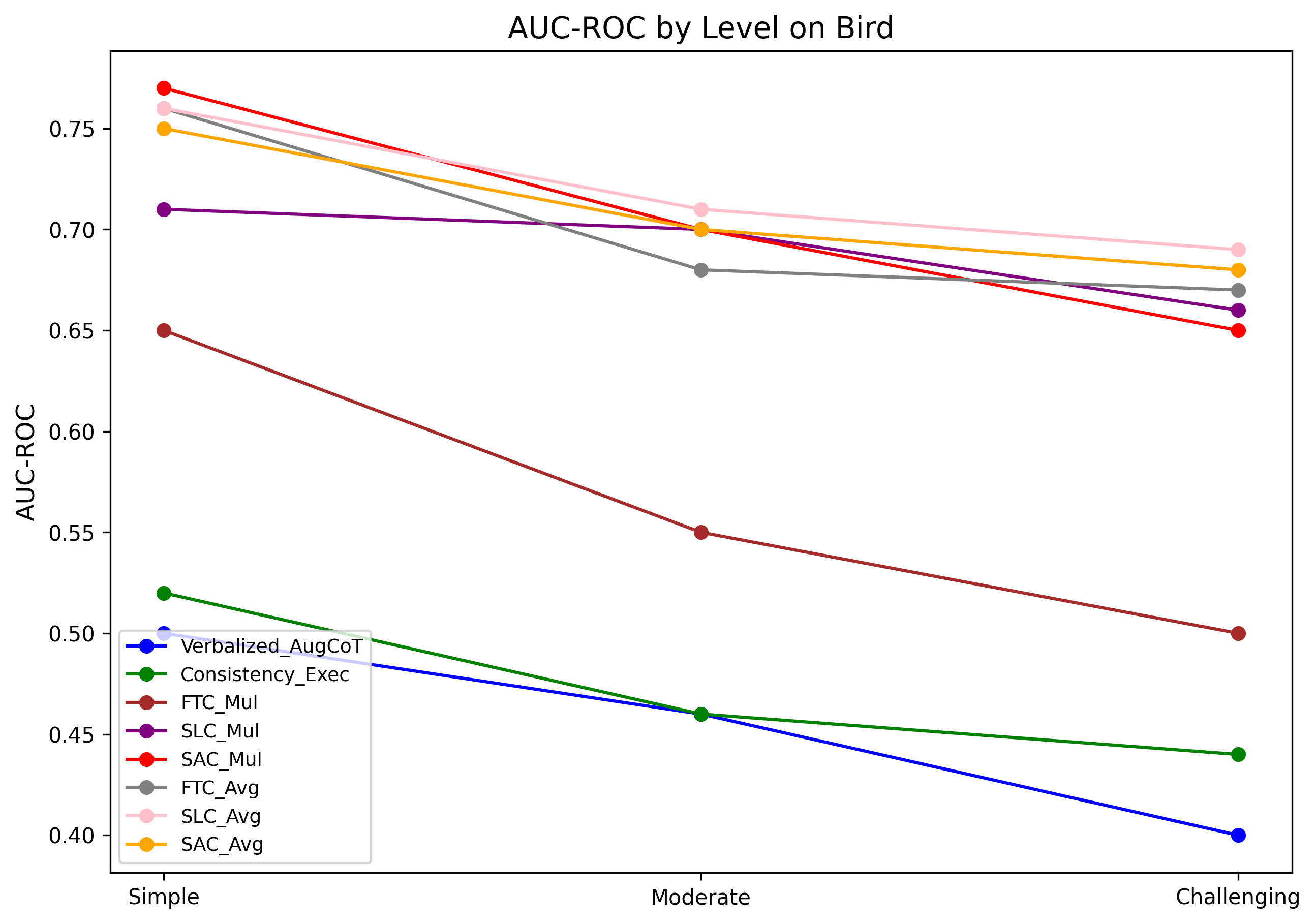}

\caption{Performance across these predefined difficulty levels in the BIRD dataset and GPT-3.5 model.}
\label{fig:Bird_level}
\end{figure}
Figures~\ref{fig:Bird_level} depict model performance across various query difficulty levels for the Bird dataset.
Challenging queries typically feature heavier schema-linked, multi-table joins, aggregations, and nested clauses. As the SQL queries become more complex, the models exhibit reduced confidence in token generation, resulting in lower confidence scores for the generated queries, as expected. 

The figure shows a clear trend of declining model performance as query difficulty progresses from simple to challenging. This decline is most pronounced for the execution-based Consistency, Augmented-COT Verbalized, and FTC-Mul methods, where the models show a steep drop in performance. In contrast, the SAC models exhibit a more gradual decline in performance, indicating a better handling of complex SQL structures due to their focus on crucial SQL tokens and schema elements. Unlike FTC-Mul, which is adversely affected by the larger number of tokens in complex queries, SAC models manage to maintain relatively higher mean scores across all levels of difficulty, underscoring their robustness in face of complexity. These trends are consistent with observations from the Spider dataset, as detailed in Figure~\ref{fig:Spider_level} in the appendix. However, it is noted that models utilizing multiple aggregation techniques and the execution-based Consistency show relatively better performance stability on the Spider dataset, indicating their effectiveness in managing simpler queries. 

\subsection{Performance Varying Query Length} 
\begin{figure}[t]
\centering
\includegraphics[width=\columnwidth,trim=10 5 10 10,clip]{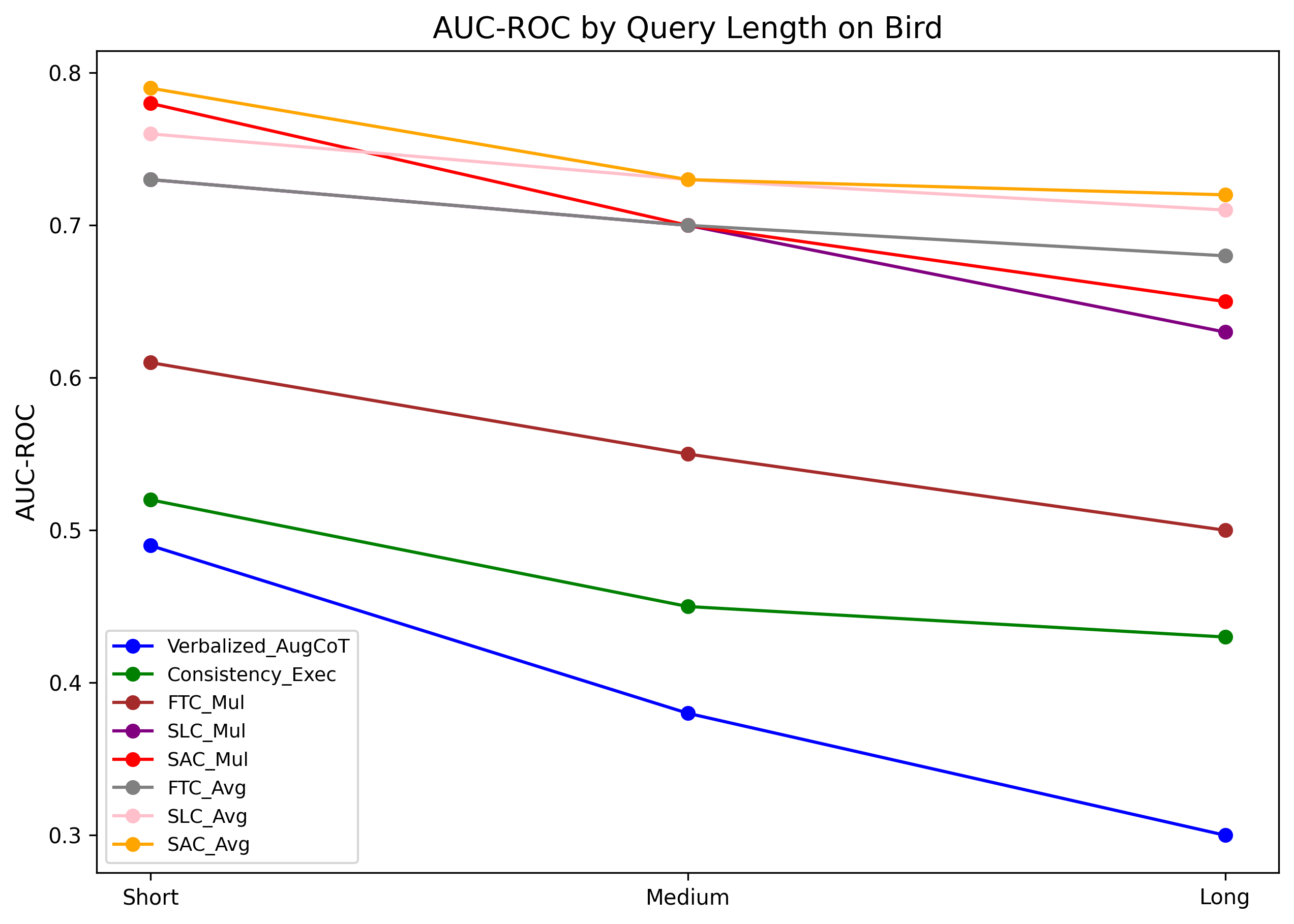}

\caption{The distribution of query lengths and model performance (GPT-3.5) on the BIRD dataset.}
\label{fig:Bird_length_category}
\end{figure}
Figure~\ref{fig:Bird_length_category} illustrates how the Black-box and White-box approaches perform as average query lengths vary within the Bird benchmark. We utilized histograms to determine appropriate binning intervals for SQL query lengths, categorizing them as Short (0-15 tokens), Medium (16-25 tokens), and Long (above 26 tokens). For Spider, the majority of queries, 5,046, were categorized as Short, indicating a prevalence of less complex or shorter queries within this dataset. Additionally, 2,749 queries were classified as Medium, and 1,844 as Long. For Bird we have a significant prevalence of longer queries, with 5,966 falling into the Long category, compared to 651 in the Medium and 597 in the Short categories. 

The length of a query often correlates with its complexity, and we observe a general trend of diminishing performance as query length increases. This is particularly noticeable in the FTC model, where longer queries exacerbate the impact of low-probability tokens, leading to significant score reductions. The SAC model, in contrast, demonstrates superior performance in handling longer queries.

The choice of aggregation method distinctly affects performance: multiplication aggregations tend to magnify penalties associated with longer queries, adversely impacting the FTC model. In contrast, average aggregation smooths out these effects and ensures more stable performance across different query lengths. This highlights that aggregation functions which normalize based on query length tend to yield better results, especially in the context of longer and more complex queries. A contrasting scenario is observed in the Spider dataset, as depicted in Figure~\ref{fig:Spider_length_category} (see Appendix). Here, multiplication aggregation functions appear more effective, largely because the queries are generally shorter compared to those in the Bird dataset.


\subsection{Performance Varying Schema Link Size} 
\begin{figure}[t]
\centering
\includegraphics[width=\columnwidth,trim=10 5 10 10,clip]{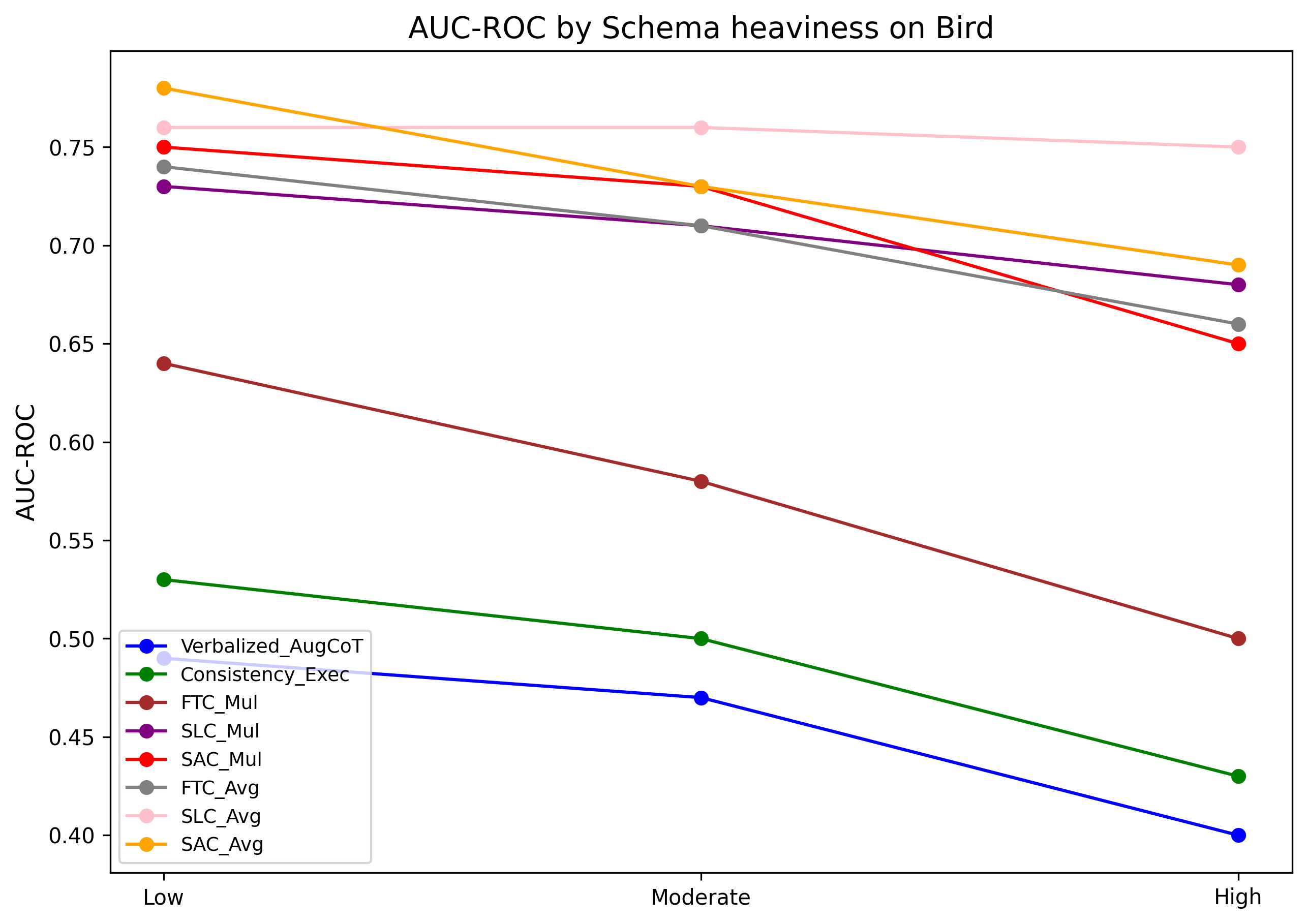}

\caption{The distribution of schema heaviness and model (GPT-3.5) performance in the BIRD dataset.}
\label{fig:Bird_schema_category}
\end{figure}
Figure~\ref{fig:Bird_schema_category} visualizes the impact of schema complexity on models performance within the Bird benchmark. Schema heaviness was categorized into three bins based on the number of schema link tokens: Low (0-5 tokens), Moderate (6-9 tokens), and High (above 10 tokens). For the Spider dataset, the majority of queries (4979) were categorized as Low, indicating a prevalence of simpler or shorter queries, followed by 4074 Moderate and 605 as High complexity queries. In contrast, the Bird dataset presents a significant prevalence of schema-heavy queries, with 4947 falling into the High category, compared to 1172 in Moderate and 1105 in the Low category. 
Schema heaviness often correlates with increased query complexity as queries with more schema links---such as a greater number of tables, columns, and values---tend to exhibit reduced confidence scores.

In environments characterized by schema-heavy queries, the SLC and SAC models consistently outperform other methods. These models prioritize schema-linked tokens, which play a pivotal role in enhancing the accuracy of confidence estimation for Text\-to\-SQL tasks. By focusing primarily on schema links, even to the exclusion of other factors, these models provide a robust measure of model confidence. This approach is particularly effective in the Bird dataset, where schema complexity is high. A similar pattern is observed in the Spider dataset, confirming the utility of this approach across different levels of schema heaviness, as shown in Figure~\ref{fig:Spider_schema_category} in the appendix.

The choice of aggregation strategy, average versus product, affects performance depending on query characteristics, such as complexity and length. As shown in Figure ~\ref{fig:Bird_level} for BIRD, SAC-Mul performs best on simple queries, but its advantage narrows as queries grow more complex, with SAC-Avg eventually outperforming it. A similar pattern is seen in Figure ~\ref{fig:Spider_level} for Spider, though the effect is less pronounced, likely due to Spider's generally shorter and simpler queries. These trends are further supported by analyses of varying query length (Figures ~\ref{fig:Bird_length_category} and ~\ref{fig:Spider_length_category}) and schema heaviness (Figures ~\ref{fig:Bird_schema_category} and ~\ref{fig:Spider_schema_category}).

\subsection{Impact of Execution Feedback}
Incorporating execution feedback into our SQL query models significantly enhances performance, particularly with SAC methods which exhibit the highest AUC scores across the Spider and Bird datasets. Notably, execution grounding substantially improves results on the more complex Bird dataset, achieving an AUC of 79.06 and an ECE of 12.15 for GPT-3.5. Detailed results and methodological comparisons are provided in the appendix.

\subsection{Impact of SQL-Specific Features}

\begin{table}[t]                  
  \centering
  \small                          
  \setlength{\tabcolsep}{5pt}     
  \begin{tabular}{lcc}
    \toprule
    \textbf{Feature Category} & \textbf{AUC} & \textbf{ECE$\downarrow$}\\
    \midrule
   SAC-Average          & 79.06  & 12.15 \\
w/o Token Exclusion  & $-5.7\%$ & $+4.2\%$ \\
w/o Case Folding     & $-3.8\%$ & $+1.3\%$ \\
w/o Order Folding    & $-1.2\%$ & $+1.1\%$ \\
w/o Synonym Folding  & $-0.5\%$ & $+0.3\%$ \\
w/o Equiv.\ Expr.    & $-0.7\%$ & $+0.4\%$ \\

    \bottomrule
  \end{tabular}
  \caption{Impact of ablating SQL-specific features on SAC-Avg (BIRD).}
  \label{tab:abl_features}
\end{table}

Table \ref{tab:abl_features} highlights the importance of SQL-specific features on the performance of SAC-Average on the Bird dataset. Excluding non-critical tokens led to the largest performance drop (-5.7\% AUC, +4.2\% ECE), showing its central role in accurate confidence estimation. Case and order folding also proved important, reducing AUC by 3.8\% and 1.2\% when removed.  Synonym and equivalent expression folding contributed more marginal but still meaningful improvements, suggesting that handling synonymous constructs and logically interchangeable expressions further refines confidence estimation. These findings confirm that incorporating SQL-specific conventions enhances the model's robustness and predictive reliability.

\section{Conclusions and Future Work}
This work presents an exploration of confidence estimation for text-to-SQL, addressing a critical gap in the application of LLMs for natural language to SQL translation. 
Our study demonstrates the superiority of white-box methods, particularly the SQL-Aware model, which excels across diverse queries due to SQL-specific adjustments and robust aggregation strategies like average aggregation. While black-box methods, such as consistency-based approaches, offer simplicity, they are constrained by latency and cost. Future work could explore hybrid methods combining the strengths of white-box and black-box approaches, alternative aggregation strategies for complex queries, and evaluations on real-world databases. 

\bibliography{ref}

\clearpage
\begin{center}
    \Large\bfseries Supplementary Material
\end{center}
\appendix

\section{Appendix}
\subsection{Prompts}

\subsubsection{Vanilla Verbalized Method}
\label{app-vanilla-verbalized}

This section provides the prompt used for Vanilla verbalized method.

\begin{figure}[h]
    \centering
    \begin{tikzpicture}
      \node[draw, rectangle, inner sep=2mm, fill=mylightblue] (rect) {
        \begin{minipage}{\linewidth}
          \texttt{You are an agent designed to answer the given question by generating a SQL Query and provide your confidence level. \\
Note that the confidence level indicates the degree of certainty you have about the SQL Query (between 0 and 100). \\
The schema of the tables are provided with three samples rows from each table. Use this information for generating the SQL Query. \\
\\
follow the below format: \\
Question: user's question \\
SQL Query: SQL query that can answer the question\\
Confidence: A score in range (0-100) which shows how confident your are about the answer \\
\\
TABLE\_SCHEMA
\\
\\
Question: QUESTION \\
SQL Query: \\
}
\end{minipage}
      };
    \end{tikzpicture}
    \caption{Template for Vanilla Verbalized Method}
    \label{fig:sqlquerygeneration_Vanilla}
  \end{figure}


\subsubsection{COT Verbalized Method}

This section provides the prompt used for chain-of-thought verbalized method.

\begin{figure}[h]
    \centering
    \begin{tikzpicture}
      \node[draw, rectangle, inner sep=2mm, fill=mylightblue] (rect) {
        \begin{minipage}{\linewidth}
          \texttt{You are an agent designed to generate a SQL query and a confidence score. \\
Confidence score indicates how correctly and accurately SQL Query answers the user's question. \\
Note that the confidence score should be between 0 and 100. \\
Use the scratchpad to evaluate the SQL Query step by step to obtain the confidence score. \\
After using the scratchpad always write your final score after Confidence:. \\
You are provided with the schema of tables and three first rows of them, user's question, and the SQL Query. \\
An example: \\
Question: How many singers are older than 20 years old? \\ 
SQL Query: SELECT COUNT(*), name FROM singer WHERE age >= 20 \\
Scratchpad: Let's think step by step, the query asks for the number of singers whose age is larger than 20. \\
The SQL Query is expected to return the name of the singers with the number of singers with age older than or equal to 20, but the question only asks for the number of singers. \\
There are two problems with the SQL Query: 1) name is redundant in the SELECT statement 2) WHERE statement should be age > 20 not age >= 20
The SQL Query has major problems so the confidence score for this SQL query should be low. \\ 
Confidence: 10 \\
\\
TABLE\_SCHEMA
\\
\\
Give me a SQL query and a confidence score for the given question. \\
Question: QUESTION \\
SQL Query:
}
\end{minipage}
      };
    \end{tikzpicture}
    \caption{Template for COT Verbalized Method}
    \label{fig:sqlquerygeneration_COT}
  \end{figure}

\subsubsection{White-box Method Prompt}

This section provides the prompt used for White-box methods.

\begin{figure}[h]
    \centering
    \begin{tikzpicture}
      \node[draw, rectangle, inner sep=2mm, fill=mylightblue] (rect) {%
        \begin{minipage}{\linewidth}
          {\ttfamily
            You are an agent designed to answer the given question by generating a SQL Query and provide your confidence level.\\
            Note that the confidence level indicates the degree of certainty you have about the SQL Query (between 0 and 100).\\
            The schema of the tables are provided with three samples rows from each table. Use this information for generating the SQL Query.\\[1ex]
            follow the below format:\\
            Question: user's question\\
            SQL Query: SQL query that can answer the question\\
            Confidence: A score in range (0-100) which shows how confident you are about the answer\\[1ex]
            TABLE\_SCHEMA\\[1ex]
            Question: QUESTION\\
            SQL Query:
          }
        \end{minipage}
      };
    \end{tikzpicture}
    \caption{Template for Whitebox Method}
    \label{fig:sqlquerygeneration_whitebox}
\end{figure}


\subsection{Additional Experiment Details}

\subsubsection{Dataset and Model Descriptions}
\label{app:datasets-llms}
\paragraph{Datasets}
The Spider dataset consists of 1,034 examples across 20 databases, and the BIRD dataset includes 1,533 question-SQL pairs from 11 databases, covering diverse domains such as healthcare, finance, and education. The BIRD dataset presents more complex SQL query challenges compared to Spider.

\paragraph{Models}
Experiments were conducted using two distinct models: GPT-3.5-turbo~\citep{brown2020language} and DeepSeek 6.7B~\citep{guo2024deepseek}. GPT-3.5-turbo, which has approximately 175 billion parameters, is renowned for its broad applicability across a variety of natural language processing tasks. It demonstrates strong generalization capabilities and robust performance in diverse linguistic contexts. DeepSeek 6.7B, with 6.7 billion parameters, is specifically optimized for tasks requiring a deep understanding of structured query language and is effective in generating complex SQL queries. This model combines advanced natural language processing techniques with specialized architectures to enhance performance on structured data interpretation.

\subsubsection{Evaluation Metrics}
\label{app:metrics}
\paragraph{Expected Calibration Error (ECE)}
The Expected Calibration Error (ECE) quantifies the calibration of our models by measuring the discrepancy between predicted probabilities and observed accuracy. Lower ECE values indicate better calibration, suggesting that the model's confidence estimates are more closely aligned with the actual likelihood of correct predictions. A value close to zero indicates optimal calibration, where the model's confidence is highly indicative of its predictive accuracy.

\paragraph{Area Under the Receiver Operating Characteristic Curve (AUC-ROC)}
The Area Under the Receiver Operating Characteristic Curve (AUC-ROC) measures the model's ability to distinguish between correctly and incorrectly generated SQL queries at various confidence score thresholds. Higher AUC-ROC values, closer to 1.0, suggest that the model is highly effective at classifying queries correctly based on its confidence scores. A value of 0.5 indicates no discriminative power, equivalent to random guessing, and values below 0.5 suggest performance worse than random guessing.

\subsubsection{Detailed Methodology for SQL-Aware Confidence Estimation}
\label{app:sql_adjustments}
In refining our SQL-aware confidence estimation, we adjust the probabilities for candidate tokens at each position, with access to all candidate lists where the sum of probabilities equals one. Key adjustments include handling case insensitivity, synonymous keywords, and managing interchangeable constructs, such as symmetric conditions and order interchangeability within clauses. For case insensitivity, we normalize SQL keywords by consolidating the probabilities of various cases (e.g., `SELECT` and `select`) into a single probability for the canonical form. For synonymous expressions like \texttt{!=} and \texttt{<>}, we combine their probabilities to reflect their semantic equivalence.

For interchangeable constructs, including conditions like `x=y` and `y=x`, and order variations within clauses such as `SELECT a, b` and `SELECT b, a`, we determine the probabilities of permutations by identifying each syntactically valid variation that the model might generate. This involves summing the individual probabilities of these variations as they appear in the candidate lists for each token position. For example, if `SELECT a, b` and `SELECT b, a` are potential outputs, we sum the probabilities assigned to each sequence by the model, thereby ensuring that our confidence measure reflects the true likelihood of either sequence being correct without being biased by the order in which tokens appear. This method leverages SQL's flexibility and enhances the model's accuracy in reflecting the semantic integrity of queries.
\newpage

\subsection{Query difficulty, Query lengths, and Schema heaviness in the Spider dataset}
\label{app:varying-diff-spider}

\begin{figure}[ht]
\centering
\includegraphics[width=\linewidth]{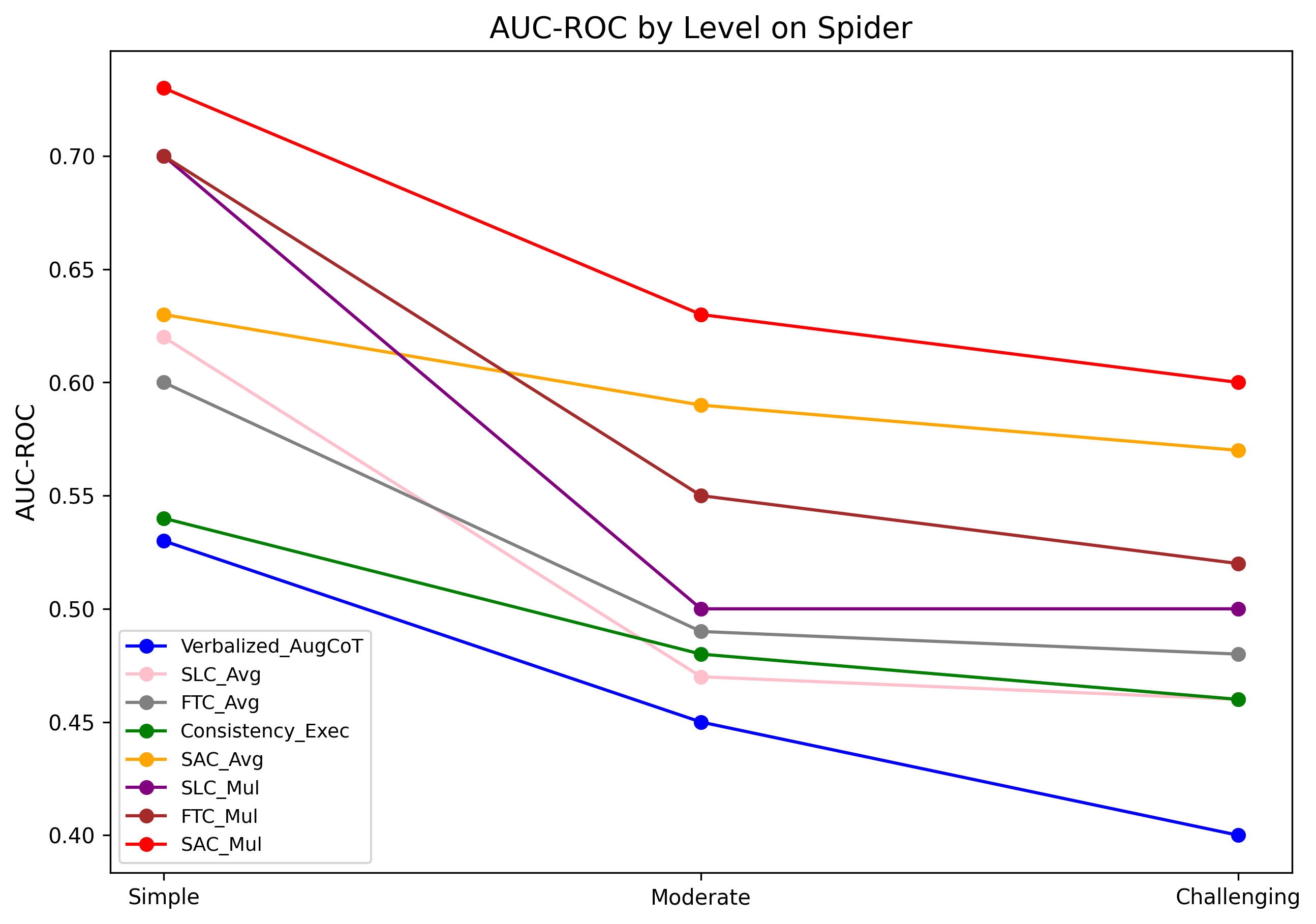}
\caption{Performance of models across varying difficulty levels in the Spider dataset, measured using the AUC-ROC metric.}
\label{fig:Spider_level}
\end{figure}

\begin{figure}[ht]
\centering
\includegraphics[width=\linewidth]{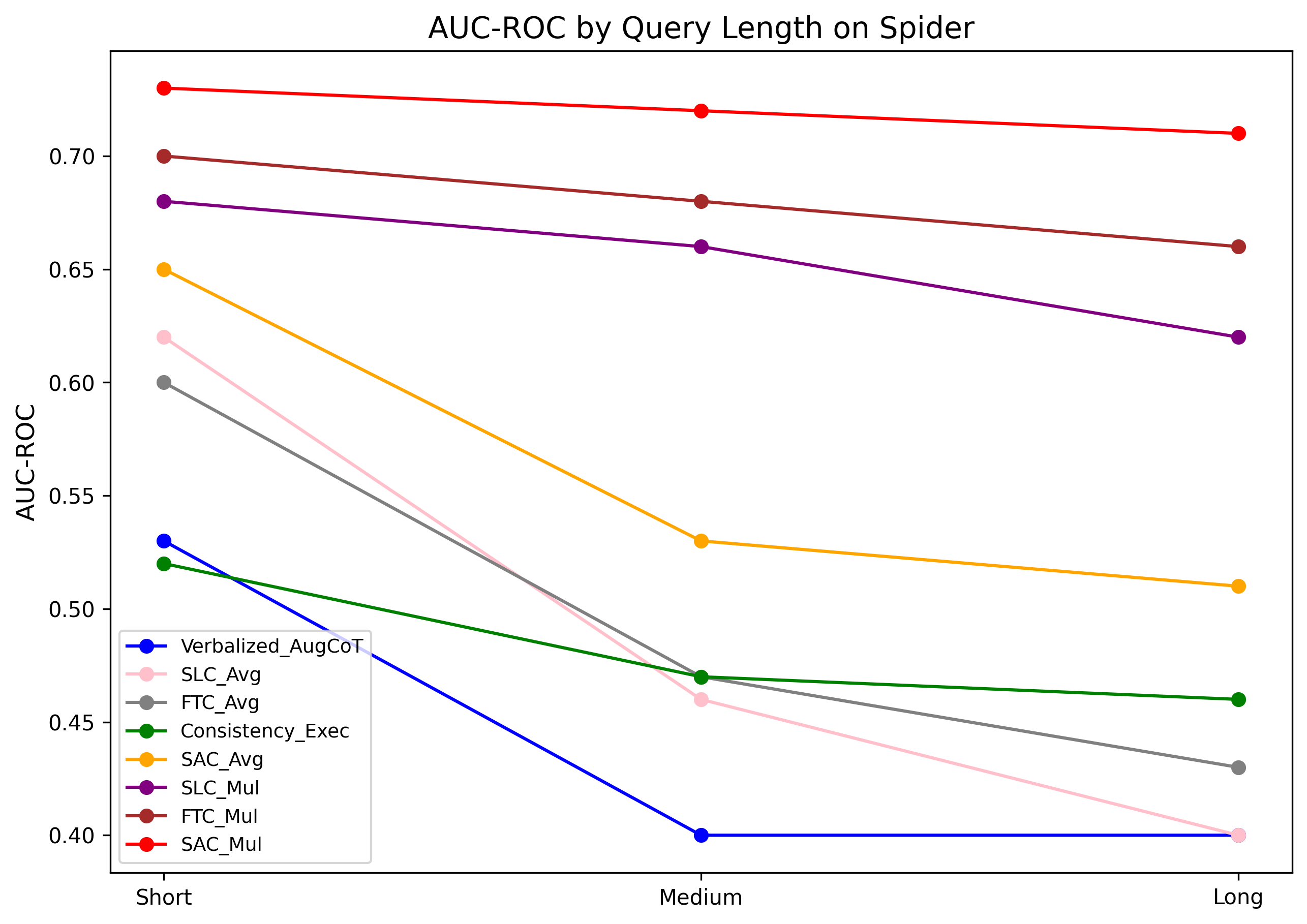}
\caption{Performance of models across varying query lengths in the Spider dataset, measured using the AUC-ROC metric.}
\label{fig:Spider_length_category}
\end{figure}

\begin{figure}[ht]
\centering
\includegraphics[width=\linewidth]{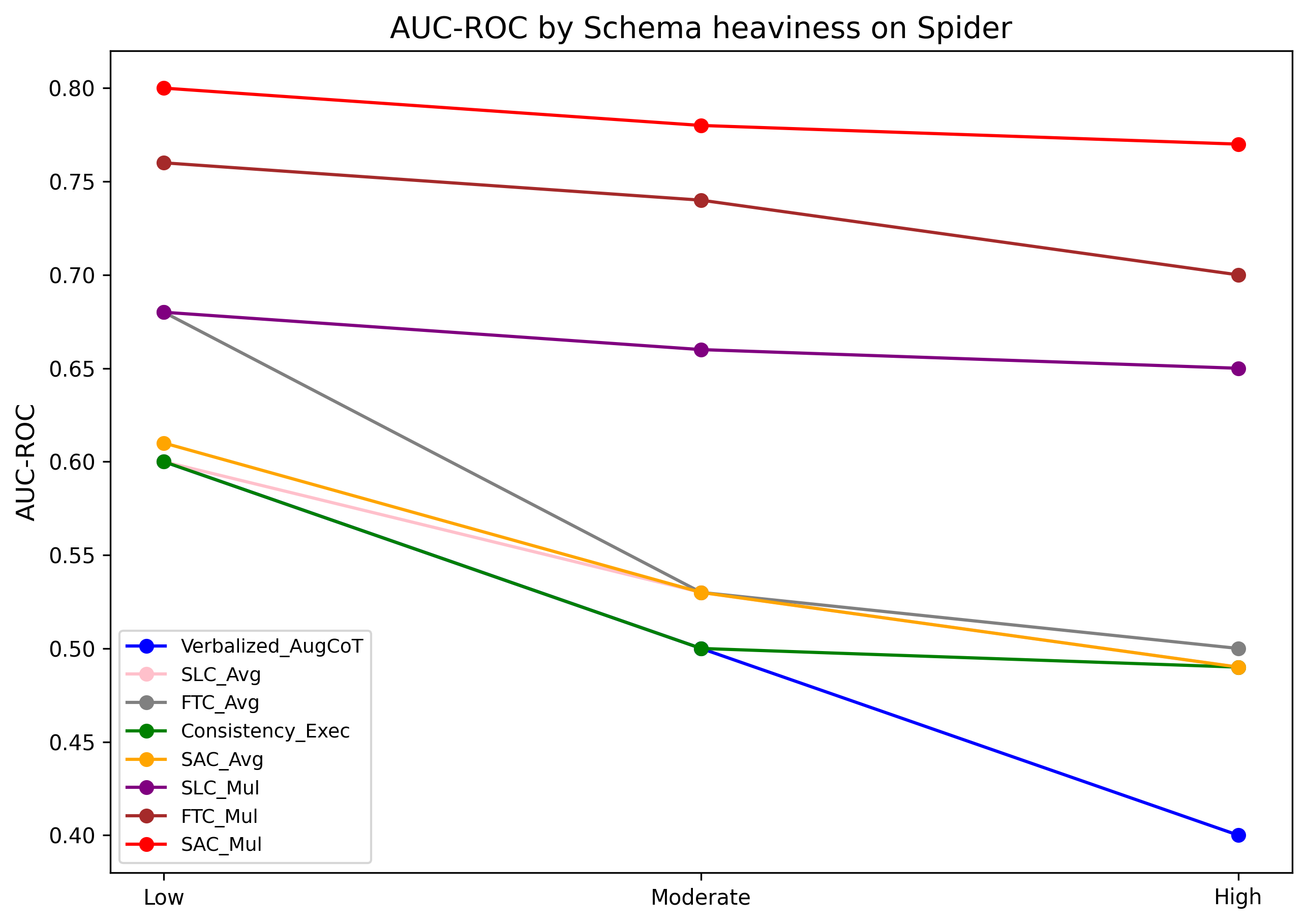}
\caption{Performance of models across varying schema heaviness in the Spider dataset, measured using the AUC-ROC metric.}
\label{fig:Spider_schema_category}
\end{figure}


\subsection{Query difficulty, Query lengths, and Schema heaviness in the Bird dataset,  measured by ECE metric.}

\label{app:varying-diff-bird}
\begin{figure}[ht]
\centering
\includegraphics[width=\linewidth]{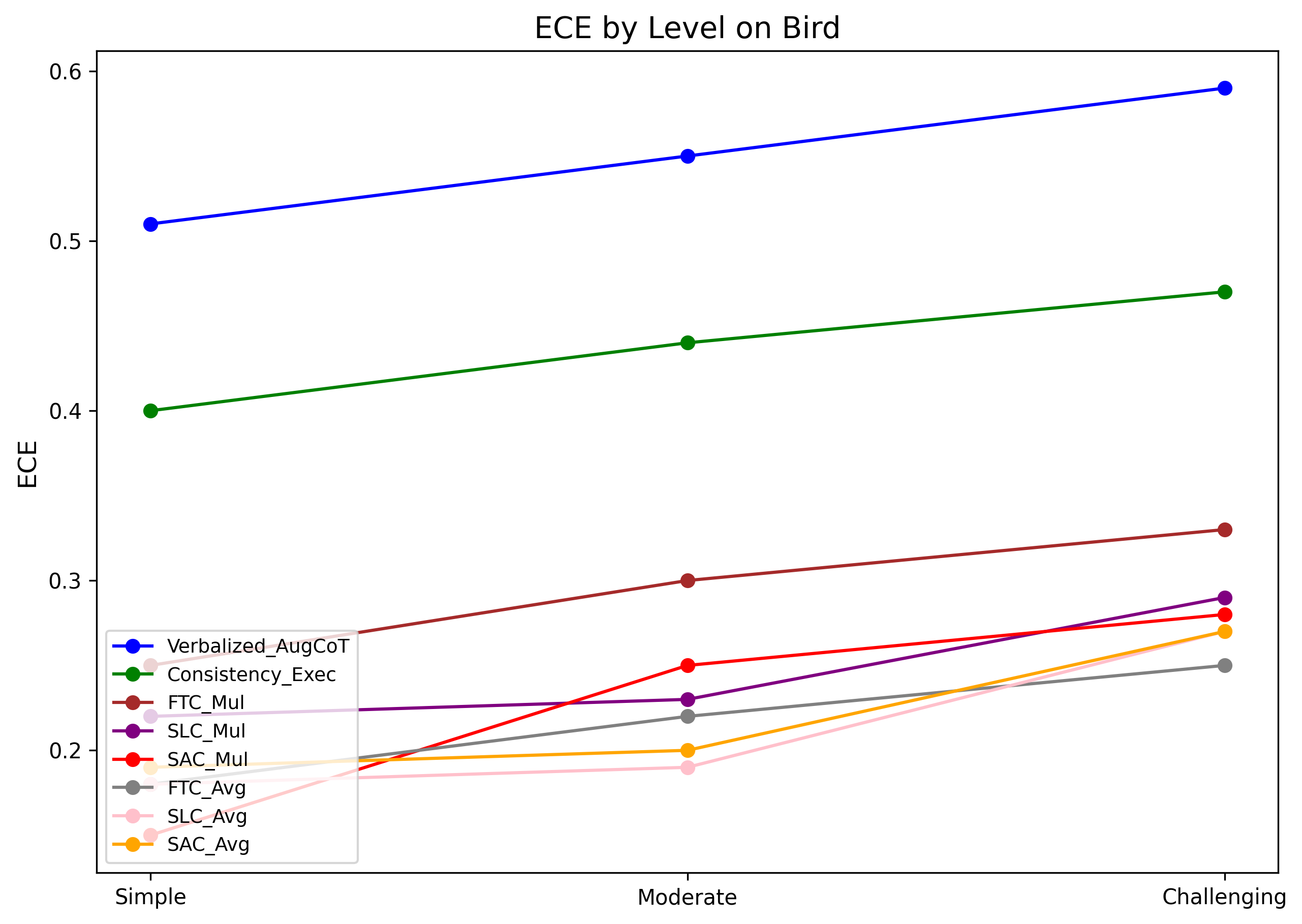}
\caption{Performance of models across varying difficulty levels in the Bird dataset, measured using the ECE metric.}
\label{fig:ece_LEVEL_bird}
\end{figure}

\begin{figure}[ht]
\centering
\includegraphics[width=\linewidth]{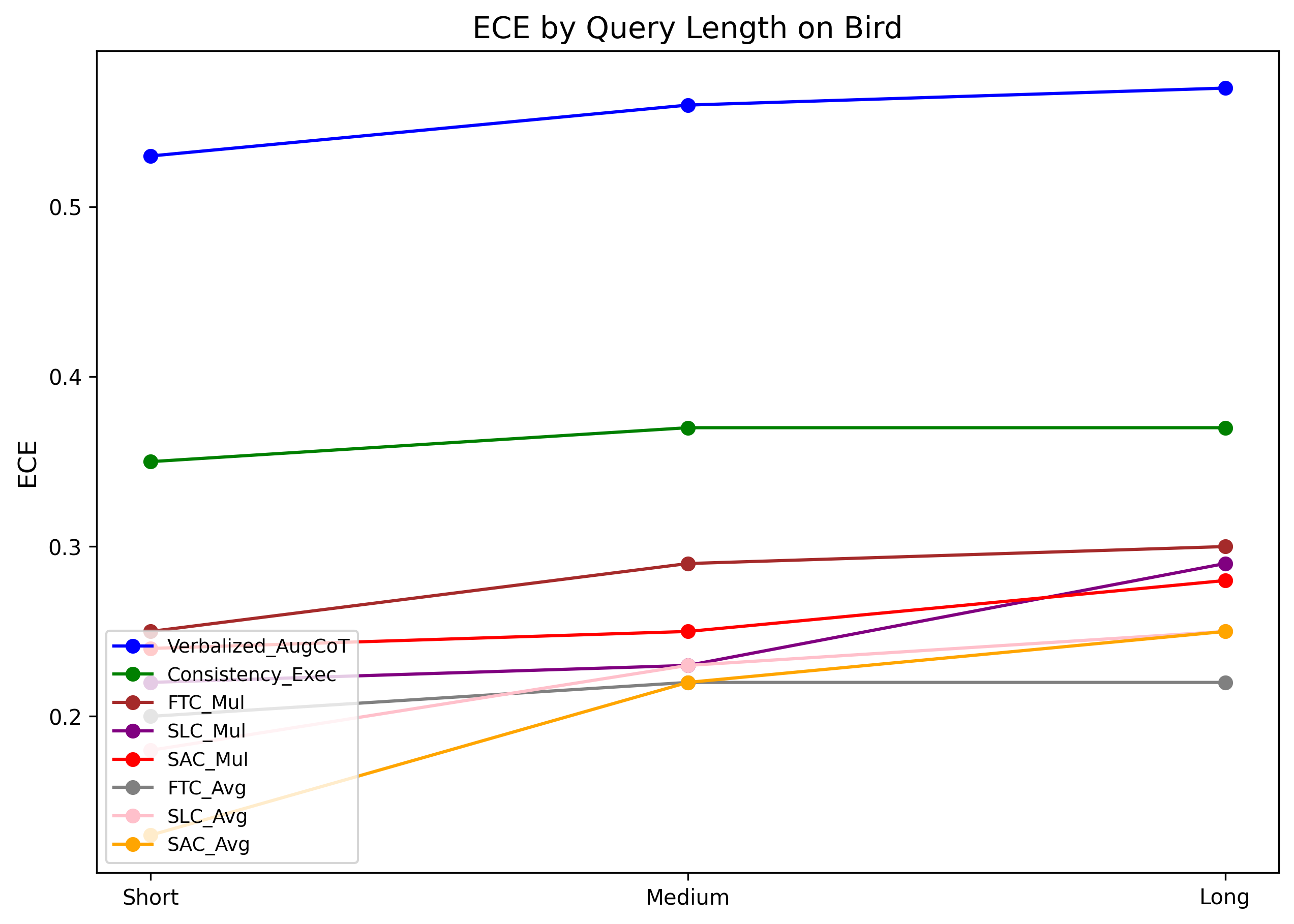}
\caption{Performance of models across varying query lengths in the Bird dataset, measured using the ECE metric.}
\label{fig:ece_QL_bird}
\end{figure}

\begin{figure}[ht]
\centering
\includegraphics[width=\linewidth]{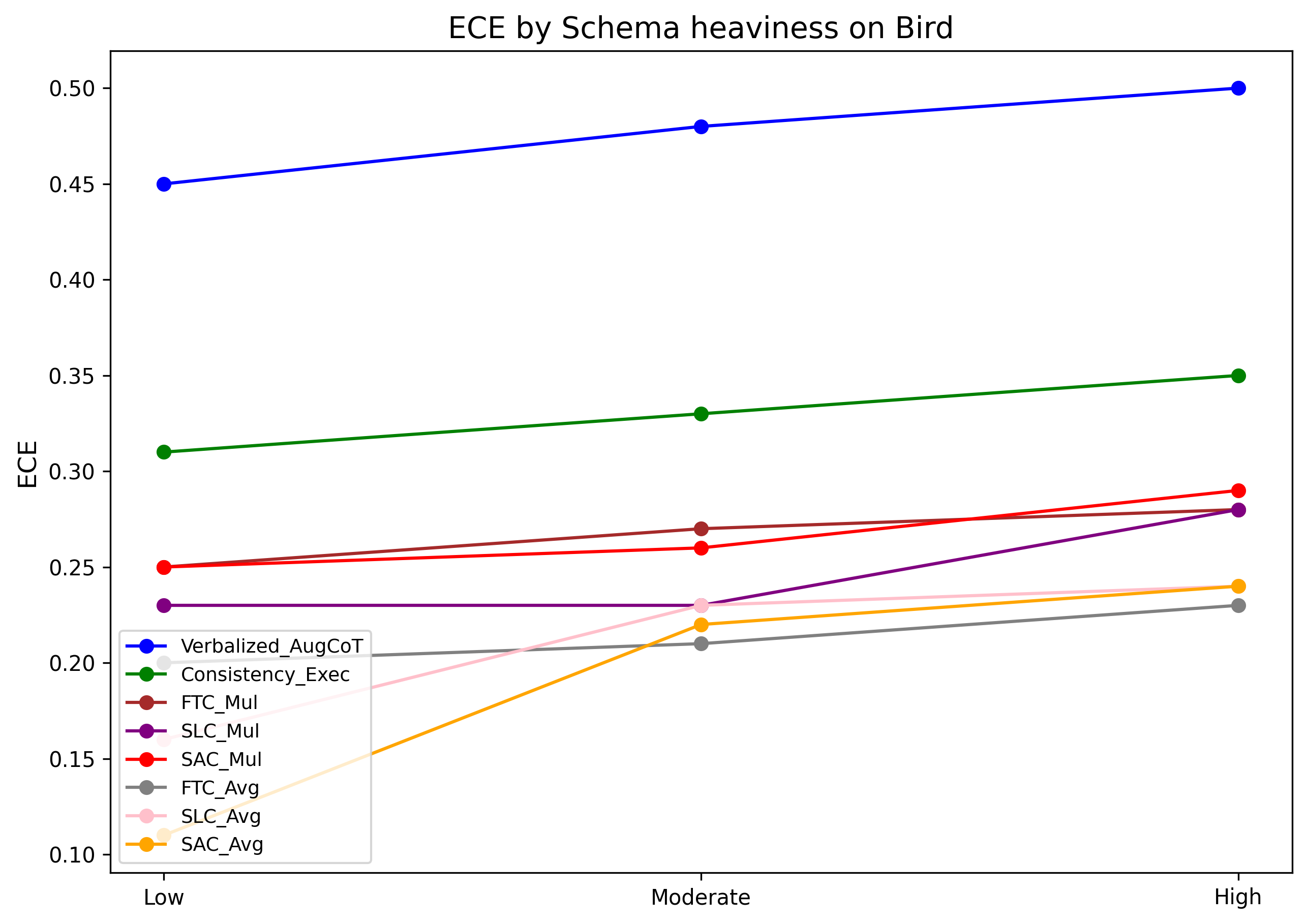}
\caption{Performance of models across varying schema heaviness in the Bird dataset, measured using the ECE metric.}
\label{fig:ece_SL_bird}
\end{figure}


\begin{table*}[!htbp]
\caption{\small Performance of our models on Spider and Bird dev sets (w/ non-executive SQLs). Abbreviations: FTC = Full Token Confidence, SLC = Schema-Linked Confidence, SAC = SQL-Aware Confidence.}
\label{table:ApproachesResults_non-executive}
\centering
\small
\renewcommand{\arraystretch}{1.2} 
\setlength{\tabcolsep}{3pt} 
\begin{tabular}{l l l c c c c}
\hline
\textbf{Dataset} & \textbf{Approach} & \textbf{Method} & \multicolumn{2}{c}{\textbf{GPT-3.5}} & \multicolumn{2}{c}{\textbf{DeepSeek}} \\
 & & & \textbf{AUC} $\uparrow$& \textbf{ECE}$\downarrow$ & \textbf{AUC}$\uparrow$ & \textbf{ECE}$\downarrow$ \\
\hline
\multirow{12}{*}{Spider} 
 & \multirow{6}{*}{Logit-based} 
 & FTC - Average & 62.78 & 22.10 & 58.12 & 24.41 \\
 & & FTC - Product & 67.57 & 19.38 & 62.18 & 20.00 \\
 & & SLC - Average & 62.21 & 23.70 & 52.81 & 29.07 \\
 & & SLC - Product & 64.89 & 22.13 & 61.11 & 22.67 \\
 & & SAC - Average & 64.33 & 23.08 & 57.79 & 27.90 \\
 & & SAC - Product & \textbf{71.14} & \textbf{16.98} & \textbf{64.81} & \textbf{19.80} \\
 \cline{2-7}
 & \multirow{6}{*}{Black-box} 
 & Consistency-based (Execution-based) & \textbf{65.01} & \textbf{19.38} & \textbf{62.85} & \textbf{20.22} \\
 & & Consistency-based (Embedding-based) & 58.22 & 24.84 & 60.24 & 23.27 \\
 & & Consistency-based (Schema-based) & 59.12 & 23.43 & 61.01 & 22.48 \\
 & & Verbalized (Vanilla) & 55.05 & 26.31 & 55.31 & 26.43 \\
 & & Verbalized (COT) & 57.44 & 25.23 & 61.07 & 23.81 \\
 & & Verbalized (Augmented COT) & 58.70 & 23.54 & 60.94 & 22.97 \\
\hline
\multirow{12}{*}{Bird} 
 & \multirow{6}{*}{Logit-based} 
 & FTC - Average & 62.97 & 21.60 & 59.67 & 24.99 \\
 & & FTC - Product & 57.25 & 34.51 & 51.11 & 33.16 \\
 & & SLC - Average & 65.77 & 21.16 & 60.51 & 22.63 \\
 & & SLC - Product & 60.99 & 26.83 & 53.78 & 28.69 \\
 & & SAC - Average & \textbf{68.13} & \textbf{15.87} & \textbf{63.84} & \textbf{18.42} \\
 & & SAC - Product & 59.65 & 23.77 & 56.27 & 25.63 \\
 \cline{2-7}
 & \multirow{6}{*}{Black-box} 
 & Consistency-based (Execution-based) & \textbf{67.01} & \textbf{28.04} & \textbf{65.98} & \textbf{28.92} \\
 & & Consistency-based (Embedding-based) & 61.01 & 31.32 & 60.66 & 34.78 \\
 & & Consistency-based (Schema-based) & 63.98 & 30.11 & 64.16 & 30.91 \\
 & & Verbalized (Vanilla) & 54.57 & 60.63 & 53.11 & 54.98 \\
 & & Verbalized (COT) & 56.02 & 54.13 & 58.04 & 51.61 \\
 & & Verbalized (Augmented COT) & 57.73 & 52.29 & 59.21 & 46.31 \\
\hline
\end{tabular}
\end{table*}

\subsection{Detailed Impact of Execution Feedback}
\label{Detailed_Impact_of_Execution_Feedback}

\begin{table*}[!htbp]
\caption{\small Performance of our models on Spider and Bird dev sets with and without execution grounding of SQL queries.}
\label{table:ABL_ApproachesResults}
\centering
\small
\renewcommand{\arraystretch}{1} 
\setlength{\tabcolsep}{3pt} 
\begin{tabular}{l l l c c c c}
\hline
\textbf{Dataset} & \textbf{Approach} & \textbf{Method} & \multicolumn{2}{c}{\textbf{GPT-3.5}} & \multicolumn{2}{c}{\textbf{DeepSeek}} \\
 & & & \textbf{AUC} $\uparrow$& \textbf{ECE}$\downarrow$ & \textbf{AUC}$\uparrow$ & \textbf{ECE}$\downarrow$ \\
\hline
\multirow{8}{*}{Spider} 
 & \multirow{4}{*}{w/o execution grounding} 
 & SAC - Average & 64.33 & 23.08 & 57.79 & 27.90 \\
 & & SAC - Product & \textbf{71.14} & \textbf{16.98} & \textbf{64.81} & \textbf{19.80} \\
 & & Consistency-based (Execution-based) & 65.01 & 19.38 & 62.85 & 20.22 \\
 & & Verbalized (Augmented COT) & 58.70 & 23.54 & 60.94 & 22.97 \\
 
 \cline{2-7}
 & \multirow{4}{*}{with execution grounding} 
 & SAC - Average & 65.01 & 23.08 & 58.12 & 27.90 \\
 & & SAC - Product & \textbf{71.66} & \textbf{16.98} & \textbf{65.26} & \textbf{19.80} \\
 & & Consistency-based (Execution-based) & 65.91 & 19.13 & 63.25 & 19.98 \\
 & & Verbalized (Augmented COT) & 58.70 & 23.54 & 60.94 & 22.97 \\
\hline
\multirow{8}{*}{Bird} 
 & \multirow{4}{*}{w/o execution grounding} 
 & SAC - Average & \textbf{68.13} & \textbf{15.87} & \textbf{63.84} & \textbf{18.42} \\
 & & SAC - Product & 59.65 & 23.77 & 56.27 & 25.63 \\
  & & Consistency-based (Execution-based) & 67.01 & 28.04 & 65.98 & 28.92 \\
 & & Verbalized (Augmented COT) & 57.73 & 52.29 & 59.21 & 46.31 \\
 \cline{2-7}
 & \multirow{4}{*}{with execution grounding} 
 & SAC - Average & \textbf{79.06} & \textbf{12.15} & \textbf{77.94} & \textbf{11.06} \\
 & & SAC - Product & 73.0 & 22.88 & 71.19 & 21.11 \\
 & & Consistency-based (Execution-based) & 67.36 & 27.14 & 66.48 & 28.34 \\
 & & Verbalized (Augmented COT) & 57.73 & 52.29 & 59.21 & 46.31 \\

\hline
\end{tabular}
\end{table*}

Table \ref{table:ABL_ApproachesResults} provides a comprehensive view of the performance variations across different models on the Spider and Bird datasets when incorporating execution feedback. The results highlight nuanced differences based on dataset complexity and the underlying approach.
%
SAC methods consistently show the highest AUC scores across both datasets, emphasizing their robust handling of SQL queries. Execution feedback leads to noticeable performance improvements for both GPT-3.5 and DeepSeek, especially in the Bird dataset. SAC Average on the Bird dataset improves dramatically with execution grounding, achieving an AUC of 79.06 for GPT-3.5 and 77.94 for DeepSeek, along with the lowest ECE scores (12.15 and 11.06, respectively). This suggests that execution-based corrections are highly effective for logit-based methods that otherwise lack direct execution feedback. The Bird dataset's intricate schema and complex query structures further amplify the impact of execution grounding. The substantial performance gains for SAC methods on this dataset underscore the critical role of execution-based grounding in enhancing model robustness for challenging query scenarios.

As expected, the execution-based consistency approach exhibits only marginal improvements with execution grounding. AUC scores for GPT-3.5 and DeepSeek improve by 0.35 and 0.50 points, respectively, on the Bird dataset, while ECE remains relatively high. This modest gain is attributed to the inherent design of the approach, which already factors in execution feedback by grouping queries with identical results.
%
The Augmented-COT Verbalized model shows no observable impact from execution grounding. AUC and ECE scores remain unchanged across both datasets. This is consistent with the model's architecture, which already includes query results (either successful output or error messages) in its input prompt. Therefore, execution feedback does not introduce additional information that could further enhance performance.

\subsection{Case Study Analysis of Models}
\label{app-case-study}

\subsubsection{Distribution of Question Difficulty and Model Accuracy}
\begin{figure}[t]
\centering
\includegraphics[width=\columnwidth,trim=10 5 10 10,clip]{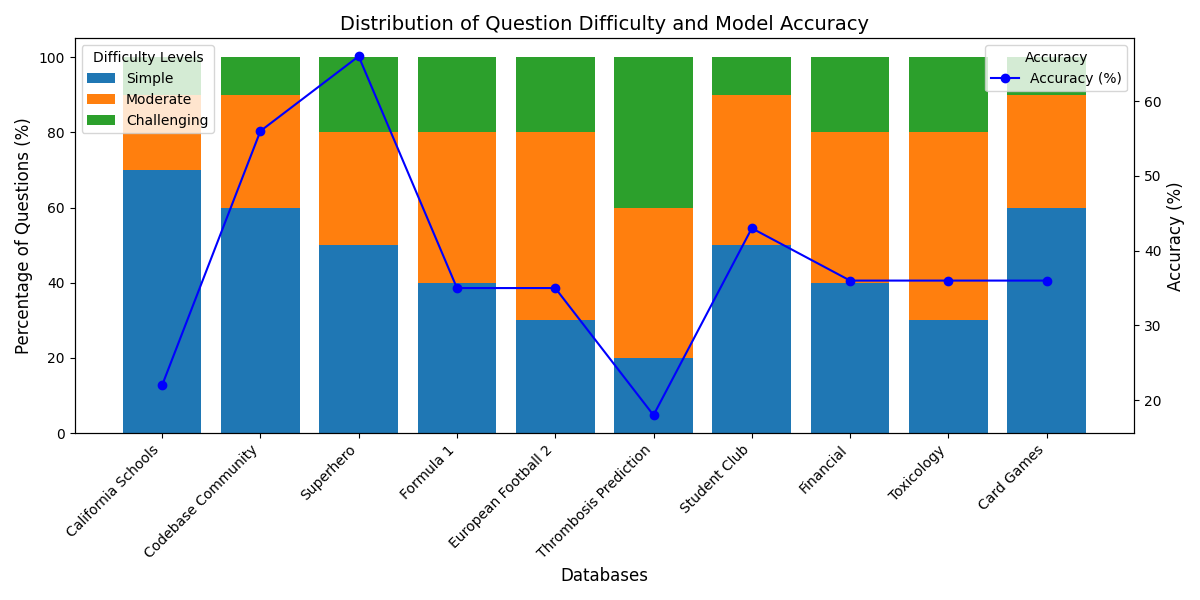}
\caption{Distribution of question difficulty levels and model accuracy across domains in the BIRD dataset.}
\label{fig:Bird_databases}
\end{figure}

Figure \ref{fig:Bird_databases} illustrates the distribution of question difficulty levels (simple, moderate, and challenging) and model accuracy across various domains in the BIRD dataset\citep{wretblad-etal-2024-understanding}. The chart highlights the relationship between query difficulty and model performance, serving as a foundation for selecting the Thrombosis Prediction and Codebase Community datasets for deeper analysis in this work.

\subsubsection{Case Study Analysis of White-box modelsfor Thrombosis Prediction and Codebase Community}

We conducted an analysis of our White-box models by examining several samples and their generated tokens from the datasets. Figure \ref{fig:Bird_databases} presents the distribution of question difficulty levels and model accuracy across databases in the BIRD dataset \citep{wretblad-etal-2024-understanding}. For this study, we selected two datasets: \textit{Thrombosis Prediction} and \textit{Codebase Community}, representing challenging and simpler datasets, respectively. We presented various examples from the databases by detailing the question, both the generated and gold SQL, the query's characteristics, its label, and our primary white-box models using the average aggregation function. Here is the schema of the databases:

\noindent
\textbf{Database Schema for Thrombosis Prediction:}
\noindent
\textbf{Table Examination}, columns = [*, ID, Examination Date, aCL IgG, aCL IgM, ANA, ANA Pattern, aCL IgA, Diagnosis, KCT, RVVT, LAC, Symptoms, Thrombosis]
\noindent
\textbf{Table Patient}, columns = [*, ID, SEX, Birthday, Description, First Date, Admission, Diagnosis]
\noindent
\textbf{Table Laboratory}, columns = [*, ID, Date, GOT, GPT, LDH, ALP, TP, ALB, UA, UN, CRE, T-BIL, T-CHO, TG, CPK, GLU, WBC, RBC, HGB, HCT, PLT, PT, APTT, FG, PIC, TAT, TAT2, U-PRO, IGG, IGA, IGM, CRP, RA, RF, C3, C4, RNP, SM, SC170, SSA, SSB, CENTROMEA, DNA, DNA-II]

\noindent
\textbf{Database Schema for Codebase Community:}
\noindent
\textbf{Table badges}, columns = [*, Id, UserId, Name, Date]
\noindent
\textbf{Table comments}, columns = [*, Id, PostId, Score, Text, CreationDate, UserId, UserDisplayName]
\noindent
\textbf{Table postHistory}, columns = [*, Id, PostHistoryTypeId, PostId, RevisionGUID, CreationDate, UserId, Text, Comment, UserDisplayName]
\noindent
\textbf{Table postLinks}, columns = [*, Id, CreationDate, PostId, RelatedPostId, LinkTypeId]
\noindent
\textbf{Table posts}, columns = [*, Id, PostTypeId, AcceptedAnswerId, CreaionDate, Score, ViewCount, Body, OwnerUserId, LasActivityDate, Title, Tags, AnswerCount, CommentCount, FavoriteCount, LastEditorUserId, LastEditDate, CommunityOwnedDate, ParentId, ClosedDate, OwnerDisplayName, LastEditorDisplayName]
\noindent
\textbf{Table tags}, columns = [*, Id, TagName, Count, ExcerptPostId, WikiPostId]
\noindent
\textbf{Table users}, columns = [*, Id, Reputation, CreationDate, DisplayName, LastAccessDate, WebsiteUrl, Location, AboutMe, Views, UpVotes, DownVotes, AccountId, Age, ProfileImageUrl]
\noindent
\textbf{Table votes}, columns = [*, Id, PostId, VoteTypeId, CreationDate, UserId, BountyAmount]



\subsubsection{Thrombosis Prediction Dataset}

\begin{enumerate}
\item \textbf{Question:} How many under-age patients were examined during the three-year period from 1990 to 1993?\\[2pt]

\textbf{Generated SQL:}
{\footnotesize%
\begin{verbatim}
SELECT COUNT(*)
  FROM Patient
  INNER JOIN Examination ON Patient.ID = Examination.ID
 WHERE (strftime('%Y','now') 
 - strftime('%Y',Birthday)) < 18
   AND (strftime('%Y','now') 
   - strftime('%Y','1993-12-31')) < 3;
\end{verbatim}}

\textbf{Gold SQL:}
{\footnotesize%
\begin{verbatim}
SELECT COUNT(T1.ID)
  FROM Patient AS T1
  INNER JOIN Examination AS T2 ON T1.ID = T2.ID
 WHERE STRFTIME('%Y',
 T2.'Examination Date') BETWEEN '1990' AND '1993'
   AND STRFTIME('%Y',T2.'Examination Date')
       - STRFTIME('%Y',T1.Birthday) < 18;
\end{verbatim}}

\textbf{Query Features:} length = 91, schema heaviness = 9, difficulty = High.\\
\textbf{Method Scores:} FTC 0.8689, SLC 0.8860, SAC 0.9836.\\
\textbf{Label:} True\\
\textbf{Explanation:} \emph{FTC} is slightly under-confident because it assigns low
probability to non-critical tokens (e.g., punctuation, numeric literals).
\emph{SLC} is also mildly under-confident, overlooking key logical tokens
(\texttt{strftime}, \texttt{AND}). \emph{SAC} focuses on schema-linked tokens,
yielding a high confidence score without over- or under-estimation.

\item \textbf{Question:} What is the disease patient \texttt{30609} was
diagnosed with? List all laboratory–test dates for this patient.\\[2pt]

\textbf{Generated SQL:}
{\footnotesize%
\begin{verbatim}
SELECT "Examination"."Examination Date",
       "Laboratory"."Date"
  FROM "Examination"
  JOIN "Patient"
    ON "Examination"."ID" = "Patient"."ID"
  JOIN "Laboratory"
    ON "Patient"."ID" = "Laboratory"."ID"
 WHERE "Patient"."ID" = 30609;
\end{verbatim}}

\textbf{Gold SQL:}
{\footnotesize%
\begin{verbatim}
SELECT T1.Diagnosis, T2.Date
  FROM Patient AS T1
  INNER JOIN Laboratory AS T2 ON T1.ID = T2.ID
 WHERE T1.ID = 30609;
\end{verbatim}}

\textbf{Query Features:} length = 88, schema heaviness = 9, difficulty = Simple.\\
\textbf{Method Scores:} FTC 0.8796, SLC 0.8326, SAC 0.7178.\\
\textbf{Label:} False\\
\textbf{Explanation:} \emph{FTC} is over-confident because its broad token
evaluation assigns high scores to non-critical items (e.g.,
\texttt{"."}, \texttt{1.000}, \texttt{"}, \texttt{0.920}).  \emph{SLC} is
better aligned but remains slightly over-confident by scoring
\texttt{JOIN} at \texttt{0.772} while still overlooking the truly
important tokens.  \emph{SAC} lowers its score appropriately, reflecting
the query’s incorrectness.

\item \textbf{Question:} For patients with albumin ${<}\,3.5$, list their
ID, sex, and diagnosis.\\[2pt]

\textbf{Generated SQL:}
{\footnotesize%
\begin{verbatim}
SELECT ID, SEX, Diagnosis
  FROM Patient
 WHERE ID IN (
       SELECT ID
         FROM Laboratory
        WHERE ALB < 3.5);
\end{verbatim}}

\textbf{Gold SQL:}
{\footnotesize%
\begin{verbatim}
SELECT DISTINCT T1.ID, T1.SEX, T1.Diagnosis
  FROM Patient AS T1
  INNER JOIN Laboratory AS T2 ON T1.ID = T2.ID
 WHERE T2.ALB < 3.5;
\end{verbatim}}

\textbf{Query Features:} length = 43, schema heaviness = 5, difficulty = Simple.\\
\textbf{Method Scores:} FTC 0.8937, SLC 0.8553, SAC 0.9568.\\
\textbf{Label:} True\\
\textbf{Explanation:} \emph{FTC} is mildly mis-calibrated: it gives high
probability to generic separators (commas, \texttt{FROM}) yet lowers the
score for a few non-critical tokens, diluting overall confidence.
\emph{SLC} shows slight under-confidence by under-weighting the filtering
predicate \texttt{IN}, which is essential for selecting the correct
records.  \emph{SAC}, focused on schema and predicate tokens, produces a
well-aligned confidence score.

\item \textbf{Question:} How many male patients, tested between 1995 – 1997
and later diagnosed with Behçet disease, did \emph{not} stay in the
hospital for treatment?\\[2pt]

\textbf{Generated SQL:}
{\footnotesize%
\begin{verbatim}
SELECT COUNT(*)
  FROM Patient
  INNER JOIN Examination ON Patient.ID = Examination.ID
 WHERE Patient.SEX = 'M'
   AND Examination.Diagnosis LIKE '%BEHCET%'
   AND Examination.'Examination Date'
       BETWEEN '1995-01-01' AND '1997-12-31'
   AND Patient.Admission = '-';
\end{verbatim}}

\textbf{Gold SQL:}
{\footnotesize%
\begin{verbatim}
SELECT COUNT(T1.ID)
  FROM Patient AS T1
  INNER JOIN Examination AS T2 ON T1.ID = T2.ID
 WHERE T2.Diagnosis = 'Behcet'
   AND T1.SEX = 'M'
   AND STRFTIME('%Y',T2.'Examination Date') 
   BETWEEN '1995' AND '1997'
   AND T1.Admission = '-';
\end{verbatim}}

\textbf{Query Features:} length = 106, schema heaviness = 11, difficulty = Challenging.\\
\textbf{Method Scores:} FTC 0.8670, SLC 0.9400, SAC 0.9727.\\
\textbf{Label:} True\\
\textbf{Explanation:} \emph{FTC} is slightly under-confident because its
broad token weighting lowers the score for items such as
\texttt{'-'} (0.717) and \texttt{\%} (0.700) that matter syntactically
but not semantically.  \emph{SLC} narrows focus to schema tokens yet
still misses nuances, giving \texttt{INNER} 0.845 and \texttt{COUNT} 0.998,
yielding a modestly lower score.  \emph{SAC} concentrates on the
diagnosis filter, date range, and admission flag, producing the most
aligned confidence.

\item \textbf{Question:} Among patients born between 1936 – 1956, how many
males have creatinine-phosphokinase (CPK) values outside the normal
range?\\[2pt]

\textbf{Generated SQL:}
{\footnotesize%
\begin{verbatim}
SELECT COUNT(*)
  FROM Patient
  INNER JOIN Laboratory ON Patient.ID = Laboratory.ID
 WHERE STRFTIME('%Y',Birthday) BETWEEN '1936' AND '1956'
   AND SEX = 'M'
   AND CPK > 250;
\end{verbatim}}

\textbf{Gold SQL:}
{\footnotesize%
\begin{verbatim}
SELECT COUNT(DISTINCT T1.ID)
  FROM Patient AS T1
  INNER JOIN Laboratory AS T2 ON T1.ID = T2.ID
 WHERE STRFTIME('%Y',T1.Birthday) BETWEEN '1936' AND '1956'
   AND T1.SEX = 'M'
   AND T2.CPK >= 250;
\end{verbatim}}

\textbf{Query Features:} length = 75, schema heaviness = 9, difficulty = Challenging.\\
\textbf{Method Scores:} FTC 0.8172, SLC 0.8794, SAC 0.9862.\\
\textbf{Label:} True\\
\textbf{Explanation:} \emph{FTC} is slightly under-confident because it
penalises low-probability tokens such as \texttt{STR} (0.411) and
numeric literals (0.449) that do not alter the query logic.  \emph{SLC}
improves by focusing on schema tokens yet still underestimates the date
range and CPK predicate.  \emph{SAC} balances schema and critical
conditions, giving the most accurate confidence.

\end{enumerate}

\subsubsection{Codebase Community Dataset}

\begin{enumerate}

\item \textbf{Question:} How many positive comments are there in the list?\\[2pt]

\textbf{Generated SQL:}
{\footnotesize%
\begin{verbatim}
SELECT COUNT(*)
  FROM comments
 WHERE Score > 60;
\end{verbatim}}

\textbf{Gold SQL:}
{\footnotesize%
\begin{verbatim}
SELECT COUNT(id)
  FROM comments
 WHERE score > 60;
\end{verbatim}}

\textbf{Query Features:} length = 13, schema heaviness = 3, difficulty = Simple.\\
\textbf{Method Scores:} FTC 0.7915, SLC 0.8300, SAC 0.9988.\\
\textbf{Label:} True\\
\textbf{Explanation:} \emph{FTC} is under-confident because it assigns low
probability to \texttt{SELECT} (0.200) and \texttt{FROM} (0.701), tokens
that the LLM often mis-select at the start of a query.  \emph{SLC}
focuses on schema elements and improves the score, yet still slightly
underestimates by overlooking \texttt{COUNT} (0.996) and \texttt{*}
(0.728).  \emph{SAC} balances all critical SQL tokens, achieving
near-perfect confidence.


\item \textbf{Question:} For the \emph{most influential} user
(highest reputation), what percentage of their posts have a score
greater than 50?\\[2pt]

\textbf{Generated SQL:}
{\footnotesize%
\begin{verbatim}
SELECT
  (SELECT COUNT(*)
     FROM posts
    WHERE OwnerUserId = (SELECT Id
                           FROM users
                          ORDER BY Reputation DESC
                          LIMIT 1)
      AND Score > 50) * 100.0
  / (SELECT COUNT(*)
       FROM posts
      WHERE OwnerUserId = (SELECT Id
                             FROM users
                            ORDER BY Reputation DESC
                            LIMIT 1)) AS Percentage;
\end{verbatim}}

\textbf{Gold SQL:}
{\footnotesize%
\begin{verbatim}
SELECT CAST(SUM(CASE WHEN T2.Score >= 50 THEN 1 ELSE 0 END)
AS REAL) * 100
       / COUNT(T1.Id)
  FROM users  AS T1
  INNER JOIN posts AS T2 ON T1.Id = T2.OwnerUserId
  INNER JOIN (SELECT MAX(Reputation) AS max_rep FROM users) AS T3
             ON T1.Reputation = T3.max_rep;
\end{verbatim}}

\textbf{Query Features:} length = 92, schema heaviness = 8, difficulty = Challenging.\\
\textbf{Method Scores:} FTC 0.9046, SLC 0.9808, SAC 0.8535.\\
\textbf{Label:} False\\
\textbf{Explanation:} \emph{FTC} is over-confident: nested sub-queries and
tokens such as \texttt{(} (0.99), \texttt{AS} (0.811) and
\texttt{LIMIT} (0.91) inflate its score despite logical flaws.
\emph{SLC} likewise over-weights schema tokens and overlooks control
tokens like \texttt{COUNT} (0.644) and \texttt{ORDER} (0.863),
whereas \emph{SAC} lowers the score appropriately by focusing on the
aggregation logic, date filters, and nested selection structure.

\item \textbf{Question:} For user \#24, how many times greater is the
\emph{number of posts} than the \emph{number of votes} they received?\\[2pt]

\textbf{Generated SQL:}
{\footnotesize%
\begin{verbatim}
SELECT COUNT(p.Id) / COUNT(v.Id)
  FROM posts  AS p
  JOIN users  AS u ON p.OwnerUserId = u.Id
  LEFT JOIN votes AS v ON p.Id = v.PostId
 WHERE u.Id = 24;
\end{verbatim}}

\textbf{Gold SQL:}
{\footnotesize%
\begin{verbatim}
SELECT CAST(COUNT(T2.Id) AS REAL) / COUNT(DISTINCT T1.Id)
  FROM votes AS T1
  INNER JOIN posts AS T2 ON T1.UserId = T2.OwnerUserId
 WHERE T1.UserId = 24;
\end{verbatim}}

\textbf{Query Features:} length = 61, schema heaviness = 4, difficulty = Moderate.\\
\textbf{Method Scores:} FTC 0.817, SLC 0.836, SAC 0.785.\\
\textbf{Label:} False\\
\textbf{Explanation:} \emph{FTC} is slightly over-confident because it
rewards aliases \texttt{u} (1.0) and \texttt{v} (0.821) that do not
affect correctness.  \emph{SLC} also overestimates by giving
\texttt{COUNT} 0.563 and \texttt{LEFT} 0.763 without accounting for the
division logic.  \emph{SAC} balances schema links and arithmetic,
yielding the most appropriate (lower) confidence.


\end{enumerate}



\bigskip

\end{document}